%% file: main.tex
\documentclass[letterpaper]{article}
\input{commands}

\title{Latte-Mix: Measuring Sentence Semantic Similarity with \\Latent Categorical Mixtures}
\author{Minghan Li\footnote{Correspondence to: Minghan Li <alexlimh23@gmail.com>.}\textsuperscript{\rm 1, 2}
, He Bai\textsuperscript{\rm 1, 3}
, Luchen Tan\textsuperscript{\rm 1}
, Kun Xiong\textsuperscript{\rm 1}
, Ming Li\textsuperscript{\rm 1, 3}
, Jimmy Lin\textsuperscript{\rm 1, 3}\\
}

\affiliations{
\textsuperscript{\rm 1}RSVP.ai, \textsuperscript{\rm 2}University of Toronto,
\textsuperscript{\rm 3}David R. Cheriton School of Computer Science, University of Waterloo
}
\makeatletter
\patchcmd{\@startsection}{\@ifstar}{\nolinenumbers\@ifstar}{}{}
\patchcmd{\@xsect}{\ignorespaces}{\linenumbers\ignorespaces}{}{}
\makeatother

\begin{document}
\maketitle
\begin{abstract}
\begin{hyphenrules}{nohyphenation}
\input{sections/abstract}

\end{hyphenrules}
\end{abstract}

\input{sections/intro}
\input{sections/method}

\input{sections/experiments}
\input{sections/related}
\input{sections/discussion}

\bibliography{references}
\clearpage
\appendix
\input{sections/appendix}

\end{document}

%% file: commands.tex
\usepackage{aaai21}  
\usepackage{times}  
\usepackage{helvet} 
\usepackage{courier}  
\usepackage[hyphens]{url}  
\usepackage{graphicx} 
\usepackage{natbib}  
\usepackage{multibib}
\usepackage{caption} 
\usepackage[utf8]{inputenc}
\usepackage[T1]{fontenc}
\usepackage[english]{babel}
\usepackage[colorlinks]{hyperref}
\usepackage{amsfonts}
\usepackage{nicefrac}
\usepackage{microtype}
\usepackage{amsmath}
\usepackage{amssymb}
\usepackage{mathtools}
\usepackage{subcaption}
\usepackage{booktabs}
\usepackage{ragged2e}
\usepackage{tikz}
\usepackage{stackengine}
\usepackage{etoolbox}
\usepackage{xspace}
\usepackage{xpatch}
\usepackage{cuted}
\usepackage{enumerate}
\usepackage{xstring}
\usepackage{setspace}
\usepackage{tabularx}
\usepackage{makecell}
\usepackage{changepage}
\usepackage{enumitem}
\usepackage{cuted}
\usepackage{titlesec}
\usepackage{cancel}
\usepackage{eqparbox}
\usepackage{bibentry}
\usepackage[hang,flushmargin]{footmisc}
\usepackage[capitalise,noabbrev,nameinlink]{cleveref}
\usepackage[useregional=numeric]{datetime2}
\usepackage[linesnumbered,ruled,noend]{algorithm2e}
\usepackage{adjustbox}

\newcites{appendix}{Appendix References}
\usepackage[switch,mathlines]{lineno}
\usepackage{lipsum}
\usepackage{etoolbox}
\usepackage{longtable}

\newtheorem{theorem}{Theorem}

\newtheorem{assumption}{Assumption}

\usetikzlibrary{positioning,arrows.meta,fit,calc,backgrounds}
\tikzset{%
node distance=2em, auto,
every node/.style={line width=0.7pt},
det/.style={draw=black, rectangle, minimum size=2.5em, inner sep=0.1ex},
lat/.style={draw=black, circle, minimum size=2.5em, inner sep=0.1ex},
obs/.style={draw=black, circle, fill=black!15, minimum size=2.5em, inner sep=0.1ex},
fac/.style={draw=black, rectangle, fill=black, minimum size=.6em, inner sep=0em},
dummy/.style={draw=none, circle, minimum size=2.5em},
plate/.style={draw=black, rounded corners, inner sep=.8em, yshift=-.7em, align=right},
box/.style={draw=black, rounded corners, inner sep=.4em, align=center},
generates/.style={->, -{Stealth[length=.6em, inset=0pt]}, line width=0.7pt},
undirected/.style={line width=0.7pt},
}

\setlist[itemize]{leftmargin=1.5em,itemsep=.1em,topsep=.1em}
\makeatletter
\renewcommand{\paragraph}[1]{\textbf{#1}\quad\@ifnextchar\par\@gobble\relax}
\makeatother
\xapptocmd\normalsize{%
\abovedisplayskip=.8em plus .2em minus .2em
\belowdisplayskip=.6em plus .1em minus .1em
\abovedisplayshortskip=.8em plus .2em minus .2em
\belowdisplayshortskip=.6em plus .1em minus .1em
}{}{}

\setcitestyle{authoryear,round,citesep={;},aysep={,},yysep={;}}
\renewcommand{\cite}[1]{\citep{#1}}

\creflabelformat{equation}{#2\textup{#1}#3}
\crefname{algocf}{Algorithm}{Algorithms}
\Crefname{algocf}{Algorithm}{Algorithms}

\definecolor{mydarkblue}{rgb}{0,0.08,0.45}
\hypersetup{%
colorlinks=true,
linkcolor=mydarkblue,
citecolor=mydarkblue,
filecolor=mydarkblue,
urlcolor=mydarkblue}

\graphicspath{{figures/}}

\DeclareMathOperator*{\argmax}{arg\,max}
\DeclareMathOperator*{\argmin}{arg\,min}
\renewcommand\d{\mathop{}\!\textnormal{\slshape d}}

\makeatletter
\newcommand{\removeParBefore}{\ifvmode\vspace*{-\baselineskip}\setlength{\parskip}{0ex}\fi}
\newcommand{\removeParAfter}{\@ifnextchar\par\@gobble\relax}

\makeatother

\DeclareDocumentCommand{\p}{ D<>{p} D<>{} r() }{
\def\content{#3}\patchcmd{\content}{|}{\;#2\vert\;}{}{}
\ensuremath{#1 #2(\content #2)}}

\DeclareDocumentCommand{\P}{ D<>{P} D<>{\big} r() }{
\def\content{#3}\patchcmd{\content}{|}{\;#2\vert\;}{}{}
\ensuremath{\operatorname{#1}#2(\content #2)}}

\DeclareDocumentCommand{\E}{ D<>{E} D<>{\big} r[] }{
\def\content{#3}\patchcmd{\content}{|}{\;#2\vert\;}{}{}
\ensuremath{\operatorname{#1}\!#2[\content #2]}}

\DeclareDocumentCommand{\D}{ D<>{D} D<>{\big} r[] }{
\def\content{#3}\patchcmd{\content}{||}{\;#2\|\;}{}{}
\ensuremath{\operatorname{#1}\!#2[\content #2]}}

\NewDocumentCommand{\Nor}{ r() }{\P<Normal>](#1)}
\NewDocumentCommand{\Cat}{ r() }{\P<Cat>](#1)}
\NewDocumentCommand{\Bin}{ r() }{\P<Bin>](#1)}
\NewDocumentCommand{\Bet}{ r() }{\P<Beta>](#1)}
\NewDocumentCommand{\Ber}{ r() }{\P<Bernoulli>(#1)}
\NewDocumentCommand{\Dir}{ r() }{\P<Dir>(#1)}

\DeclareDocumentCommand{\KL}{ D<>{\big} r[] }{\D<KL><#1>[#2]}
\DeclareDocumentCommand{\H}{ D<>{\big} r[] }{\E<H><#1>[#2]}
\DeclareDocumentCommand{\I}{ D<>{\big} r[] }{\E<I><#1>[#2]}

%% file: sections/abstract.tex
Measuring sentence semantic similarity using pre-trained language models such as BERT generally yields unsatisfactory zero-shot performance, and one main reason is ineffective token aggregation methods such as mean pooling.
In this paper, we demonstrate under a Bayesian framework that distance between primitive statistics such as the mean of word embeddings are fundamentally flawed for capturing sentence-level semantic similarity.
To remedy this issue, we propose to learn a categorical variational autoencoder (VAE) based on off-the-shelf pre-trained language models. 
We theoretically prove that measuring the distance between the latent categorical mixtures, namely \textit{Latte-Mix}, can better reflect the true sentence semantic similarity. 
In addition, our Bayesian framework provides explanations for why models finetuned on labelled sentence pairs have better zero-shot performance. We also empirically demonstrate that these finetuned models could be further improved by Latte-Mix.
Our method not only yields the state-of-the-art zero-shot performance on semantic similarity datasets such as STS, but also enjoy the benefits of fast training and having small memory footprints.

%% file: sections/intro.tex
\section{Introduction}
\label{sec:intro}
The performance of pre-trained language models can be greatly improved on various NLP tasks after being finetuned on domain-specific labelled data \citep{peters2018elmo,devlin2018bert,liu2019roberta, yang2019xlnet,radford2019gpt2}.
For sentence semantics similarity tasks, although it has been found that simple pooling methods such as mean pooling could yield effective sentence representations from the pre-trained language models for classification \citep{shen2018hierpooling}, the zero-shot performance of using the distance between these representations as similarity measurement are rather poor before finetuning on task-specific data.
One solution proposed by methods like Sentence-BERT \citep{reimers2019sbert} is to further finetune the pre-trained models on labelled but out-of-domain sentence pairs.
However, in the absence of labelled data, similarity measurement based on pre-trained language models devolves into naive token aggregation whose performance can be even worse than static word embeddings \citep{reimers2019sbert}.


In this paper, we analyze the problem of mean pooling under a Bayesian framework and prove that the distance between primitive statistics, such as the mean of word embeddings, can not fully capture the sentence-level similarity.
To remedy this issue, we propose a novel approach to learn a token-wise categorical variational autoencoder (VAE) \citep{kingma2013vae,jang2016catvae} based on off-the-shelf pre-trained language models whose outputs are used as inputs and targets for the VAE. 
We theoretically show that better similarity measurements are achieved by computing the distance between the latent categorical mixtures, namely \textit{Latte-Mix}, where matching the components in latent mixtures can be seen as automatic alignment of word semantics. 
The Bayesian framework also explains why finetuning pre-trained models on labelled sentence pairs improves similarity measurement.
Empirically, our variational approach substantially improves the zero-shot performance of pre-trained language models without any finetuning and task-specific labelled data. 

To clarify, the goal of this work is not to learn better sentence representations but rather to improve the token aggregation technique for pre-trained language models.
Therefore, previous methods that improve sentence representations are orthogonal directions to this paper and will not be discussed \citep{kiros2015skipthought,conneau2017infersent,logeswaran2018quickthought,subramanian2018gensent,cer2018use,ethayarajh2018usif}.  
Furthermore, as our goal is to provide better measurement for sentence similarity, classification tasks such as SentEval \citep{conneau2018senteval} are not suitable for evaluation; it is not logically sound to use the latent distribution as features and aggregating the latent samples will degenerate to primitive pooling strategies again.
Therefore, we focus on the zero-shot performance of our method where distance between the latent distributions could be used as similarity measurement.

While finetuning on labelled data naturally yields higher performance than our reported results, the proposed metthod, to the best of our knowledge, represents the state of the art in zero-shot semantic similarity performance.


Our contributions are as follows: First, we develop a theoretical framework that formalizes the sentence semantics similarity measurement from a Bayesian viewpoint. Second, we analyze the problem of primitive token aggregation strategies such as mean pooling and prove that better similarity estimation could be achieved by measuring the distance between word semantics distributions. Lastly, based on the theoretical results, we propose a token aggregation method called Latte-Mix by learning a Categorical VAE and validate its zero-shot effectiveness on various textual semantics similarity tasks.

%% file: sections/method.tex
\section{Methodology}
\label{sec:method}
In this section, we use a common token aggregation strategy---mean pooling---as an example to show that primitive statistics of word embeddings (e.g., mean and max) are fundamentally flawed for measuring sentence semantics similarity under a Bayesian framework, while a better regime is to use the distance between the distributions of word semantics given sentences to approximate the true similarity measurement.


We begin by formalizing the measurement of sentence semantics similarity under a Bayesian framework. 
The compositionality of language implies that the meaning of a sentence can be derived from its parts, which means that semantics can be hierarchically composed \citep{emerson2020goals}. 
One way to realize compositionality is to assume a sentence semantics space distinct from the word semantics space.
Let random variable $X$ denote the word token and $U$ denote its semantics.
We use random variable $M$ for sentence-level semantics and further assume a conditional distribution $p(u\mid m)$ of word semantics $u \in U$ given the sentence semantics $m \in M$.
\begin{assumption}[Hierarchical Semantics Assumption]\label{thm:compos_assumption}
A sentence represented by a sequence of word embeddings $\{u^{(1)}, u^{(2)}\dots,u^{(k)}\}$ could be seen as i.i.d. samples drawn from a conditional semantics distribution $p(u \mid m)$, where $u$ is the word semantics and $m$ is the sentence semantics.
\end{assumption}
The conditional probability $p(u\mid m)$ could be written as
\begin{linenomath}\begin{align*}
p(u\mid m) &= \int p(u\mid x_1, x_2,\dots,x_n)p(x_1, x_2,\dots,x_n\mid m) \d x\\ 
&= \mathbb{E}_{p(x_1, x_2,\dots,x_n\mid m)}[p(u\mid x_1, x_2,\dots,x_n)],    
\end{align*}\end{linenomath}
where $p(u\mid x_1, x_2,\dots,x_n)$ corresponds to the word embedding model and $p(x_1, x_2,\dots,x_n \mid m) =\prod_t p(x_t \mid m, x_{t-1},\dots,x_1)$ is a conditional language model of human speech that outputs word token $x$ given a sentence's meaning $m$ and historical contexts. For simplicity, we will use $p(u\mid x)$ to denote both the contextual and non-contextual word embedding models in the rest of the paper.

One may argue that Assumption \ref{thm:compos_assumption} does not consider the syntactic structure or word order information of sentences. 
However, the syntactic information could also be embedded in the word vector $u$ and become parts of the semantics, which has been practiced by tree-structured models \citep{hochreiter1997lstm,tai2015treelstm}, adding positional embedding \citep{vaswani2017transformer} and POS-tag information in the word vectors.

\input{figures/mean_vs_dist/mean_vs_dist}

\subsection{Issues with Mean Pooling}

Consider two sentences $s_1$ and $s_2$ with semantics $m_1$ and $m_2$, represented by two sequences of word semantics vectors $\{u^{(i)}_1\}$ and $\{u^{(i)}_2\}$ sampled i.i.d. from two conditional semantics distribution model $p(u\mid m_1)$ and $p(u\mid m_2)$, respectively. 
The distance between their mean vectors $\bar u_1$ and $\bar u_2$ under the distance metric $d$ is given by
\begin{linenomath}\begin{align}\label{eq: mean_mc}
d\left (\bar u_1, \bar u_2\right ) &= d \left (\frac{1}{K_1}\sum_i u_1^{(i)}, \frac{1}{K_2}\sum_i u_2^{(i)}\right ) \nonumber\\ 
&\approx d \left (\mathbb{E}_{p(u\mid m_1)}[u], \mathbb{E}_{p(u\mid m_2)}[u] \right ). \end{align}\end{linenomath}

where $K_1$, $K_2$ are sequence lengths and Monte-Carlo approximation is made at the last line of Eq.~\eqref{eq: mean_mc}. 
If $m_1$ and $m_2$ are identical, $d(\mathbb{E}[u_1], \mathbb{E}[u_2])$ will be zero. However, the converse is not always true and formally established as follows.

\begin{theorem}\label{thm:mean_thm}
Consider two sequences of word semantics vectors $\{u^{(i)}_1\}_{i=1}^{K_1}$ and $\{u^{(i)}_2\}_{i=1}^{K_2}$ with sequence lengths $K_1$ and $K_2$, sampled i.i.d. from two conditional semantics distributions $p(u|m_1)$ and $p(u|m_2)$, respectively.

(1) $m_1 = m_2 \Rightarrow \mathbb{E}_{p(u\mid m_1)}[u] = \mathbb{E}_{p(u\mid m_2)}[u]$,

for every $m_1 \in M$ and $m_2 \in M$.

(2) $\mathbb{E}_{p(u\mid m_1)}[u] = \mathbb{E}_{p(u\mid m_2)}[u] \nRightarrow m_1 = m_2$,

for some $m_1 \in M$ and $m_2 \in M$.
\end{theorem}
The proof is provided in Appendix \ref{thm:mean_proof} and illustrated in Fig. \ref{fig:mean}. Theorem \ref{thm:mean_thm} indicates that measuring the distance
between the mean vectors does not guarantee to find the most similar sentences. Similar statements could also be made for other token aggregation strategies such as max pooling, with proofs similar to \ref{thm:mean_proof}. 
The conclusion here is that primitive statistics (e.g., mean and max) from the distribution $p(u\mid m)$ are not faithful measurement of sentence semantics $m$.
 
\subsection{Latent Semantics Mixtures}
To address the problem of pooling, we need to access the conditional semantics distribution $p(u\mid m)=\mathbb{E}_{p(x\mid m)}[p(u\mid x)]$ and essentially the word semantics distribution $p(u\mid x)$. Unfortunately, most pre-trained models do not provide this distribution directly, but rather a deterministic function $u = f(x)$. In this work, we leverage a VAE to approximate a latent semantics distribution 
based on the pretrained word embeddings $u$, as the VAE provides both a principled theoretical framework and an easy-to-implement algorithm in practice. 

First, we introduce a latent variable $Z$ which represents the latent semantics given the original semantics $U$. It could also be seen as an invertible transformation of the word semantics $U$. The latent semantics distribution given the original word embedding $u$ is defined as $p(z\mid u)$. Instead of calculating the distance between the expectations of word vectors $u$, we compute the distance between the distributions of latent semantics $z$. The distribution of latent semantics $z$ given the sentence semantics $m$ is 
\begin{linenomath}\begin{align}\label{eq:latent_dist}
    p(z\mid m) &= \int p(z\mid u)p(u\mid m) \d u \nonumber\\
    &= \mathbb{E}_{p(u\mid m)}[p(z\mid u)]\nonumber\\
    &\approx \frac{1}{K}\sum_i p(z\mid u^{(i)}),
\end{align}\end{linenomath}
where $K$ is the sequence length of $\{u^{(i)}\}$. Eq.~\eqref{eq:latent_dist} show that the latent semantics distribution $p(z\mid m)$ could be approximated by a mixture distribution $1/K\ast\sum_i p(z\mid u^{(i)})$ given the original word semantics $u$ in a sentence. We use a random variable $P$ to denote the space of latent distributions $p(z\mid m)$ given the sentence semantics $m$ and $D$ to denote the distance between distributions for the following theorem:

\begin{theorem}\label{thm:dist_thm}
Consider two sequences of latent semantics vectors $\{z^{(i)}_1\}_{i=1}^{K_1}$ and $\{z^{(i)}_2\}_{i=1}^{K_2}$ with sequence lengths $K_1$ and $K_2$. Suppose the following conditions are satisfied:

(1) $\{z^{(i)}_1\}_{i=1}^{K_1}$ and $\{z^{(i)}_2\}_{i=1}^{K_2}$ are sampled i.i.d. from $p(z\mid m_1)$ \\and $p(z\mid m_2)$, respectively,

(2) $p_{Z\mid M}$ is an invertible mapping from sentence semantics $M$ to the space of latent distributions $P$.

Then we have,
\begin{linenomath}
$$m_1 = m_2 \Leftrightarrow D(p(z\mid m_1), p(z\mid m_2)) = 0,$$
for every $m_1 \in M$ and $m_2 \in M$.
\end{linenomath}
\end{theorem}
which holds true for all distance metrics $D$. The proof is provided in Appendix \ref{thm:dist_proof} and illustrated in Fig. \ref{fig:dist}. From Theorem \ref{thm:dist_thm} we know that $m_1 = m_2$ if and only if $D(p(z\mid m_1), p(z\mid m_2)) = 0$. If the latent semantics distribution $p(z\mid m) \approx 1/K\ast\sum_i p(z\mid u^{(i)})$ is approximated by a mixture of Categorical distribution, then it will give rise to our method, \textit{Latte-Mix}. More reliable measurement for sentence semantic similarity could be achieved as similar word semantics distributions would come from similar sentence semantics.
Critically, this avoids the issues that we described above with simple aggregation techniques like mean pooling.

\subsection{Amortized Variational Inference} \label{subsec:vae}
The latent semantics distribution $p(z\mid u)$ mentioned in the previous section could be computed with the simple Bayes' rule: $p(z\mid u) = p(z, u)/p(u)$, where $p(u) = \int p(z, u) \d z$. However, it's generally intractable to estimate the marginal density $p(u)$ in high-dimensional space using Monte-Carlo approximation. Instead, with expressive function approximators such as neural networks \citep{goodfellow2016deep}, a variational distribution $q_{\phi}(z\mid u)$ could be used to approximate the true posterior $p(z\mid u)$ by minimizing the negative evidence lower bound (ELBO), given by
\begin{linenomath}\begin{align}\label{eq: elbo}
    \mathcal{L}(\theta, \phi; u) = \beta \ast D_{KL}[q_{\phi}(z\mid u)\mid \mid p(z)] -\nonumber\\ 
    \mathbb{E}_{q_{\phi}(z\mid u)}[\log p_{\theta}(u\mid z)], \beta > 0,
\end{align}\end{linenomath}
where $p(z)$ is some prior distribution such as $\mathcal{N}(0, 1)$ and $D_{KL}$ stands for the Kullback–Leibler divergence. $q_{\phi}(z\mid u)$ is an encoder that amortizes the cost of inferring the latent variable $z$ given each word embedding $u$ \citep{kingma2013vae,rezende2014vae}. $p_{\theta}(u\mid z)$ is a decoder that reconstructs the input word embedding $u$ from the latent variable $z$. Specifically, $\beta$ needs to be 1 for Eq. \eqref{eq: elbo} to become a legitimate negative ELBO. However, it has been shown that in practice it is beneficial to anneal $\beta$ during training \cite{fu2019cycanneal,he2019linearanneal}. 

\subsection{Choices of Reparameterization}
A common technique for obtaining differentiable and stochastic samples from $q_{\phi}(z\mid u)$ is called \textit{reparameterization} \citep{kingma2013vae}. It is often possible to express the random variable $z$ as a deterministic function $z = g_{\phi}(\epsilon, u)$, where $\epsilon$ is sampled from an independent marginal $p(\epsilon)$. A common choice for $q_{\phi}(z\mid u)$ is a Normal distribution $\mathcal{N}(z;\mu_{\phi} (u), \sigma_{\phi} (u))$. 
However, in our case, the Normal distribution might not be the best candidate for several reasons:\ (1) The Jason-Shannon divergence \citep{endres2003js} between two Gaussian mixtures does not have a closed-form solution, and Monte-Carlo method is computationally expensive for high-dimensional approximation; (2) To store the latent distribution we have to keep track of both mean and standard deviation of each token, making the space complexity increased by $\mathcal{O}(K)$ compared to the mean pooling strategy, where $K$ is the longest sentence length in a dataset.
Therefore, in this paper, we propose to use a Categorical VAE to learn the latent semantics distribution which is packed in a compact finite-dimensional vector. However, we still report the results of using the closed-form $\ell_2$ distance for the Normal VAE in Section \ref{sec:experiments:vae_hyper} for comparison.

Therefore, we choose Gumbel-Softmax \citep{jang2016catvae,maddison2016concrete} as the latent variational distribution for the Categorical VAE. 
The softmax function $\pi$ is used as a continuous, differentiable approximation to $\argmax$ which generates latent samples $z$. The $i$th dimension of $z$ could be expressed as follows:
\begin{linenomath}\begin{align}\label{eq: gumbel}
    z_i = \frac{\exp((\log(\pi_i) + g_i)/\tau )}{\sum_j \exp((\log(\pi_j ) + g_j )/\tau)},
\end{align}\end{linenomath}
where $g_1\dots g_k$ are sampled i.i.d from Gumbel$(0, 1)$\footnote{The Gumbel(0, 1) distribution can be obtained from inverse transform sampling $g = -\log(-\log(u))$ where $u \sim$ Uniform(0, 1) \citep{jang2016catvae}}. As the temperature $\tau$ approaches 0, samples from the Gumbel-Softmax distribution become one-hot and the distribution becomes identical to a categorical distribution.

\subsection{Bias in Measuring Latent Semantics}\label{sec:method:bias}
Theoretically speaking, if we have a perfect latent distribution model $p(z\mid m)$, we could accurately measure the semantic similarity between sentences according to Theorem \ref{thm:dist_thm}. However, approximation to some key entities has to be made in order to implement a practical algorithm. There are two sources of approximation bias that could not be avoided even with Monte-Carlo: the bias in the word embedding model $p_{\eta}(u\mid x)$ and the variational distribution $q_{\phi}(z\mid u)$.

As we generally do not have access to the true word semantics, we will have to choose off-the-shelf word embedding models pre-trained on large amounts of corpus \cite{mikolov2013word2vec,devlin2018bert}. Generally, these models are biased by the corpus they learned from and the optimization objectives. Therefore, the bias in the word embeddings can greatly affect the measurement on sentence semantics similarity. 

Another factor that affects semantic similarity measurement is the variational distribution $q_{\phi}(z\mid u)$. This bias is harder to evaluate as we have access to neither the true posterior $p(z\mid u)$ nor its samples. Moreover, a lower negative ELBO in Eq. \eqref{eq: elbo} does not necessarily mean a better approximation. A typical example is the KL vanishing problem \citep{chen2016lossyvae,higgins2016betavae}. The KL term in Eq. \eqref{eq: elbo} can go to 0 as the variational posterior $q_{\phi}(z|u)$ collapses to the prior $p(z)$, resulting in completely malfunctional latent code. To prevent this problem, different regularization schemes \citep{fu2019cycanneal,he2019linearanneal,kingma2016freenats} have been proposed. We explain our implementation and training schedule for the VAE model in Section \ref{sec:results} and Appendix \ref{sec:vae_hyper_all}.

\subsection{The Role of Finetuning}\label{sec:method:finetune}

Directly averaging over word embeddings from a pre-trained model such as BERT does not yield satisfactory zero-shot performance on textual semantics search \citep{reimers2019sbert,conneau2017infersent,cer2018use}. 
In fact, \citet{reimers2019sbert} found that averaging BERT's contextual embeddings is actually worse than averaging static GloVe embeddings!
To improve performance of pre-trained models, finetuning on labelled sentence pairs such as the SNLI dataset \citep{bowman2015snli} have been proposed \citep{reimers2019sbert,conneau2017infersent,cer2018use}. 
For example, methods like Sentence-BERT would first extract the mean word embeddings $\bar u_1$ and $\bar u_2$ from a sentence pair and concatenate them with other features such as the absolute difference $|\bar u_1-\bar u_2|$ to predict the labels using the siamese structure. 
Empirical results have shown that finetuning pre-trained models on such datasets yields better zero-shot performance on textual semantic similarity tasks.
Intuitively, we would of course expect this to be true, but here we provide a principled explanation using information theory.

First, the sentence relationship labels, such as contradiction and entailment, already convey the similarity information about the sentence pairs. The labels could be seen as being computed from a human score function $f(m_1, m_2)$, where $m_1$ and $m_2$ are the true semantics of the sentence pair. 
Being able to predict $f(m_1, m_2)$ from features such as $|\bar u_1 - \bar u_2|$ actually increases the mutual information between $f(m_1, m_2)$ and the distance metric $d(\bar u_1, \bar u_2)$ in Eq. \eqref{eq: mean_mc}. Let $F$ denote the random variable of the human score function $f$ and $D$ denote the random variable of $\ell_2$ distance between the mean word vectors. The mutual information $I$ between $F$ and $D$ could be expressed as
\begin{linenomath}\begin{align}\label{eq: mi}
    I(F;D) &= H(F) - H(F \mid D)\nonumber \\
           &\geq -H(F\mid D)\nonumber \\
           &= \mathbb{E}_{p(f\mid \ell_2(\bar u_1, \bar u_2))}[\ln p(f\mid \ell_2(\bar u_1, \bar u_2))] \nonumber\\
           &= \mathbb{E}_{p(f\mid |\bar u_1 - \bar u_2|)}[\ln p(f\mid |\bar u_1 - \bar u_2|)].
\end{align}\end{linenomath}
The last equality holds if the weight matrix of the linear classifier given the features $|\bar u_1 - \bar u_2|$ is initialized with a Normal distribution $\mathcal{N}(0,\sigma^2)$. Further proof is provided in Appendix \ref{thm:mi_proof}. Maximizing the last log-probability in Eq. \eqref{eq: mi}, which is the objective for finetuning on labelled sentence pairs given the feature $|\bar u_1 - \bar u_2|$, lower-bounds the mutual information between the human score function $f(m_1, m_2)$ and the metric $d(\bar u_1, \bar u_2)$. Although $d(\bar u_1, \bar u_2)$ is a flawed metric, the mutual information will be maximized if and only if $H(F|D)=0$ such that $f(m_1, m_2)$ could be consistently determined by $d(\bar u_1, \bar u_2)$. Therefore, the measurement $d(\bar u_1, \bar u_2)$ becomes more conformed with $f(m_1, m_2)$, resulting in better performance in textual semantics search tasks. In fact, \citet{reimers2019sbert} also found that $|\bar u_1 - \bar u_2|$ is an important feature for finetuning, without which the zero-shot performance on semantics search tasks will drop significantly. This result is consistent with our explanation that finetuning on labelled sentence pairs helps improve the distance metric.

In addition, in Section \ref{sec:results} we empirically show that matching the latent semantics mixtures further improves the performance of these finetuned models like Sentence-BERT with mean pooling. However, the improvement is not theoretically guaranteed as the original flawed distance metric could have been perfected by finetuning on the labelled data. 


%% file: figures/mean_vs_dist/mean_vs_dist.tex
\begin{figure}[t!]
\begin{subfigure}[t]{.5\linewidth}
\centering
\includegraphics[width=1\textwidth]{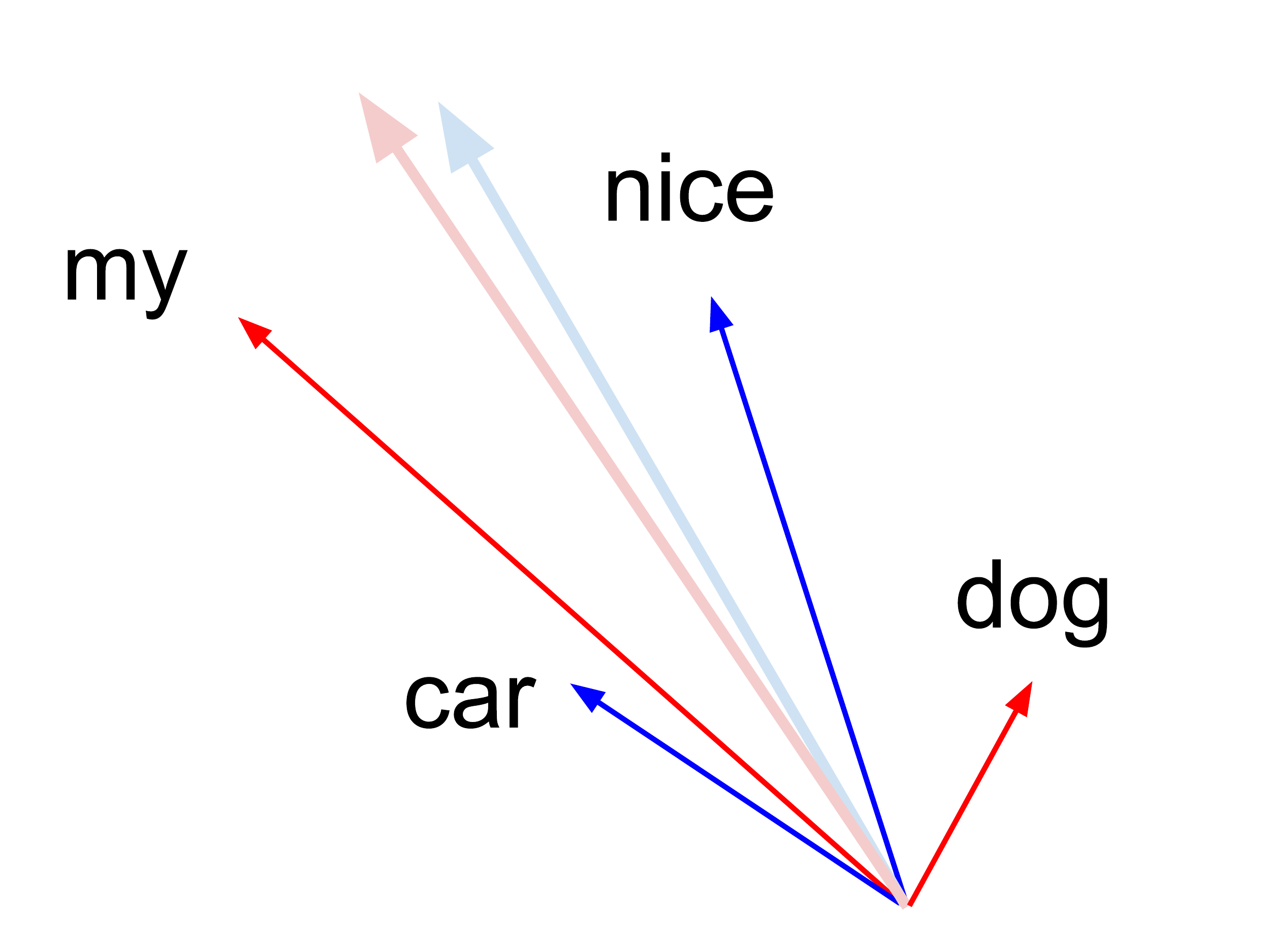}
\caption{Mean pooling}
\label{fig:mean}
\end{subfigure}%
\begin{subfigure}[t]{.5\linewidth}
\centering
\includegraphics[width=1\textwidth]{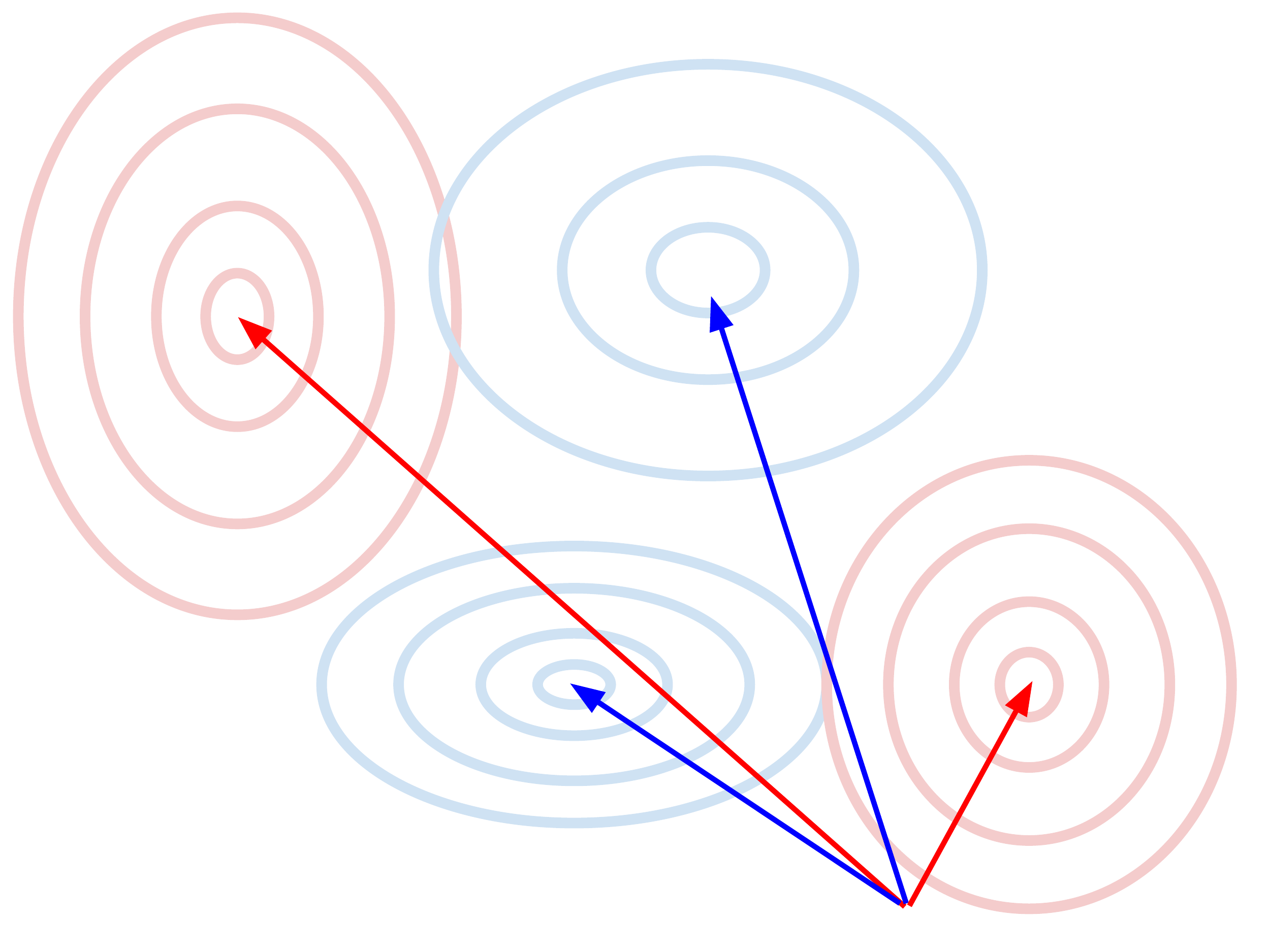}
\caption{Mixture distribution}
\label{fig:dist}
\end{subfigure}%
\caption{Comparison of two token aggregation strategies. (a) Mean pooling: similar mean vectors might come from tokens with completely different semantics. (b) Mixture distributions: measuring distance between different mixtures yields a more meaningful word semantics alignment regime.}
\label{fig:mean_vs_dist}
\end{figure}

%% file: sections/experiments.tex
\section{Experimental Setup}
\label{sec:experimental-setup}
We first conduct experiments to verify the effectiveness of the proposed method, \textit{Latte-Mix}, based on pre-trained word embeddings compared to other primitive pooling strategies such as mean/max pooling in Section \ref{sec:experiments:main}. To validate our hypothesis about the bias in sentence similarity measurement in Section \ref{sec:method:bias}, we compare different implementations of the VAE model in Section \ref{sec:experiments:vae_hyper} and use different layers of BERT as word embeddings in Section \ref{sec:experiments:bert_layers}. Lastly, we apply our method on finetuned models such as Sentence-BERT \citep{reimers2019sbert} and discover further performance gains. Complete results on  STS2012-2017\citep{agirre2012semeval,agirre2013semeval,agirre2014semeval,agirre2015semeval,agirre2016semeval,cer2017semeval} and word embedding benchmark \citep{jastrzebski2017web} are reported in Appendix \ref{sec:vae_hyper_all}, \ref{sec:sts_all} and \ref{sec:web_all}.



We use a similar experimental setup as in Sentence-BERT \citep{reimers2019sbert} where we apply Latte-Mix on off-the-shelf pre-trained language models and train it on the SNLI dataset \textit{without} using any labels. We evaluate the zero-shot performance of Latte-Mix and other baselines on Semantic Textual Similarity (STS) tasks using the Spearman's rank correlation \citep{reimers2016spearman} between the gold labels and distance of sentence embeddings/distributions.
We use cosine distance for baseline pooling strategies and measure distance between the latent mixtures with metrics like the Jason-Shannon divergence, $\ell_2$ distance and even cosine distance. Although cosine distance is not a valid metric for distributions, it is still be workable when the distribution is categorical, and sometimes outperforms the JS divergence and $\ell_2$ distance. The implementation details and hyperparameters are described in Appendix \ref{sec:appendix}.


\section{Results and Discussion}\label{sec:results}

We organize our results around four main research questions: (1) the token aggregation strategy; (2) the variational distribution $q_\phi(z|u)$; (3) the word embedding model $p_{\eta}(u|x)$; (4) whether finetuning on labelled but not task-specific sentence pairs, each described in a section below.

\input{tables/mean_vs_dist_base}
\subsection{Does Latte-Mix perform better than other pooling strategies?}\label{sec:experiments:main}

Results in Tbl. \ref{tbl:sts_base} show that measuring the distance between latent semantics distributions could substantially improve the performance compared to other pooling methods.  We also use GloVe \citep{pennington2014glove} as a strong baseline for comparing performance between contextual and non-contextual models.
We find that pre-trained language models with smaller capacity such as BERT-Base and RoBERTa-Base outperforms models with larger capacity on most datasets. This is not surprising as larger pre-trained language models is not guaranteed to provide better word embedding than the smaller models before finetuning \citep{ethayarajh2019bertemb,mickus2019assessbert}. Due to space limit, The implementation details, training procedure, and the complete results are described in Appendix \ref{sec:sts_all}.

\input{tables/vae_hyper_base}

\subsection{How does the implementation of VAE affect similarity measurement?} \label{sec:experiments:vae_hyper}
In this section, we compare different choices for latent distributions and evaluate the effect of the number of VAE's encoder layers. We mainly compare the Categorical VAE, the Normal VAE, and an autoencoder model \citep{goodfellow2016deep} which acts as a baseline. 
We use a small temperature in Eq. \eqref{eq: gumbel} for the Categorical VAE, as a small temperature could highlight each mode of $p(z|u^{(i)})$ in the latent mixture $1/K\ast\sum_i p_{\theta}(z\mid u^{(i)})$, which is more beneficial for aligning the distributions. More details about hyperparameters and results are reported in Appendix \ref{sec:vae_hyper_all}.

Results in Tbl. \ref{tbl:vae_base} show that 
the Categorical VAE outperforms the other models using the same number of encoder layers. Also, the poor performance of the autoencoder suggests that the performance gain of the VAE model is not obtained from just learning lower dimensional representations. For the same variational distribution, the performance drops as the number of encoder layers increases, where models with 1-layer encoder performs the best. As mentioned in Section \ref{sec:method:bias}, using an expressive encoder could exacerbate the KL vanishing problem as the encoder quickly collapses to the prior distribution. Therefore, using a limited-capacity encoder could provide certain regularization on the latent distribution. For the Normal VAE, as the distance between Gaussian mixtures can not be measured by the cosine distance, we use the closed-form $\ell_2$ distance as the metric in Tbl. \ref{tbl:vae_base}.

\subsection{Does using better word embedding models improve semantics search?}\label{sec:experiments:bert_layers}
The quality of the word embedding model greatly affects semantics similarity measurement for sentences. Word embeddings such as GloVe \citep{pennington2014glove} and Skip-Gram \citep{mikolov2013word2vec} can usually be evaluated on tasks such as word analogy and concept categorization. However, it is difficult to evaluate contextual word embeddings of pre-trained language models as the representations for the same word are different according to the contexts. \citet{ethayarajh2019bertemb} propose to take the first principal component of a pre-trained language model's contextualized representations in a given layer and evaluate them as static word embedding. In this section, we extract the first principle components from different layers of BERT using the STS data and evaluate them on the on the word embedding benchmark (WEB) \citep{jastrzebski2017web} as well as on the STSbenchmark dataset \citep{cer2017semeval}. 
\input{figures/layers/layers_wss}

Fig. \ref{fig:layers_wss} shows that the word embeddings from lower layers of BERT have better performance than their upper layers in a word analogy task MSR \citep{jastrzebski2017web}, which is consistent with the results from previous work \citep{ethayarajh2019bertemb}.
The results in Fig. \ref{fig:layers_sts} show that as the number of layers increases, the zero-shot performance of BERT's different layers on sentence semantics similarity is also decreasing, which verifies our hypothesis that using better word embedding models could also improve the performance in sentence semantics search. In addition, we also notice that measuring distance between latent mixtures from the VAE can greatly improve the performance lower bound compared to mean pooling, as performance gap between different layers of the pre-trained model are reduced using the VAE model in Fig. \ref{fig:layers_sts}. Notice that it is not logically sound to directly compare the word embedding between BERT and static embeddings like GloVe, as the the first principle component can not fully represent the contextual embeddings.

\input{figures/layers/layers}

\subsection{Can models finetuned on labelled sentence pairs be further improved by Latte-Mix?}
In Section \ref{sec:method:finetune} we show that finetuning on labelled sentence pairs could improve a flawed metric $d$ for similarity measurment. To verify this, we retrieve the token-wise outputs before mean pooling of the finetuned models such as Sentence-BERT \citep{reimers2019sbert} and train a VAE model on top of it the same way as in Section \ref{sec:experiments:main}. Note that what me meant by ``finetuned'' here is that the model is only trained on out-of-domain datasets such as SNLI without accessing the in-domain data such as STS before testing. Tbl. \ref{tbl:sts_sbert_base} shows that measuring the distance between latent mixtures from the 
VAE substantially improves the performance of the finetuned models. \input{tables/sbert_mean_vs_dist_base} However, unlike the results in Tbl. \ref{tbl:sts_base}, for finetuned models such as Sentence-BERT, we observe that the Jason-Shannon divergence has better performance than cosine similarity, and models with larger capacity tend to perform better. 
This is not surprising as models with larger capacity tend to perform better on downstream tasks after finetuning \citep{devlin2018bert,liu2019roberta}. 
Nevertheless, we verify that although finetuning on labelled sentence pairs already improves the 
pre-trained model's zero-shot performance, it is still beneficial to apply Latte-Mix for better measurement on sentence semantics similarity.


%% file: tables/mean_vs_dist_base.tex
\begin{table}[!t]
  \centering
  {
  \begin{tabular}{l@{\hskip .2in}l@{\hskip 0.15in}l@{\hskip 0.15in}}
\toprule
    Methods &SICK-R &STSb \\\midrule
    GloVe + Max + Cosine & 51.48  & 54.32  \\
    GloVe + Mean + Cosine & 55.38  & 61.54  \\
    GloVe + Latte-Mix + Cosine & \underline{56.49} & \textbf{65.84}* \\\hline
    BERT-Base + \texttt{[CLS]} + Cosine   &42.42  &20.29 \\
    BERT-Base + Max + Cosine    &57.98  &50.94  \\
    BERT-Base + Mean + Cosine   & 58.22  & 47.29 \\
    BERT-Base + Latte-Mix + Cosine & \textbf{61.26}* & \underline{56.82} \\\hline
    BERT-Large + \texttt{[CLS]} + Cosine &43.43   &26.75   \\
    BERT-Large + Max + Cosine &51.91   &47.18   \\
    BERT-Large + Mean + Cosine & 53.85  & 47.00  \\
    BERT-Large + Latte-Mix + Cosine  & \underline{56.71} & \underline{56.37}  \\
    \bottomrule
  \end{tabular}}
  \caption{Zero-shot spearman's rank correlation $\rho\times 100$ between the negative distance of sentence embeddings/distributions and the gold labels on SICK-R and STSb datasets. Outputs of BERT's last layer are used as inputs for the VAE. STSb: STSbenchmark, SICK-R: SICK relatedness dataset. Cosine: Cosine similarity. Mean: Mean pooling. Max: Max pooling. \texttt{[CLS]}: the embedding for the \texttt{[CLS]} token. Latte-Mix: Latent mixtures from the Categorical VAE model.} 
  \label{tbl:sts_base}
\end{table}

%% file: tables/vae_hyper_base.tex
\begin{table}[!t]
  \centering
  {
  \begin{tabular}{l@{\hskip .2in}l@{\hskip 0.15in}l@{\hskip 0.15in}}
\toprule
    Methods &SICK-R &STSb \\\midrule
    BERT-Base + Mean + Cosine   & 58.22  & 47.29 \\\hline
    AE (1-layer) + Mean + Cosine &\underline{53.90} &\underline{37.18}  \\
    AE (2-layer) + Mean + Cosine &53.82 &36.89  \\
    AE (3-layer) + Mean + Cosine &52.93 &36.79  \\\hline
    Normal-VAE (1-layer) + $\ell_2$ &\underline{55.91} &\underline{50.84} \\
    Normal-VAE (2-layer) + $\ell_2$ &54.85 &44.30  \\
    Normal-VAE (3-layer) + $\ell_2$ &53.76 &40.30  \\\hline
    Latte-Mix (1-layer) + Cosine & \textbf{61.62}* & \textbf{56.82}*   \\
    Latte-Mix (2-layer) + Cosine &55.96 &45.22   \\
    Latte-Mix (3-layer) + Cosine &55.86 &41.49   \\
    \bottomrule
  \end{tabular}
  }
  \caption{Zero-shot spearman's rank correlation $\rho \times 100$ on SICK-R and STSb datasets for the Autoencoder (AE), Categorical VAE (Latte-Mix), and Normal VAE (Normal-VAE). Outputs of the BERT-Base's last layer are used as the VAE's inputs. (N-layer): Number of the VAE's encoder layers.}
  \label{tbl:vae_base}
\end{table}

%% file: figures/layers/layers_wss.tex
\begin{figure}[!t]
\centering
\includegraphics[width=.8\linewidth]{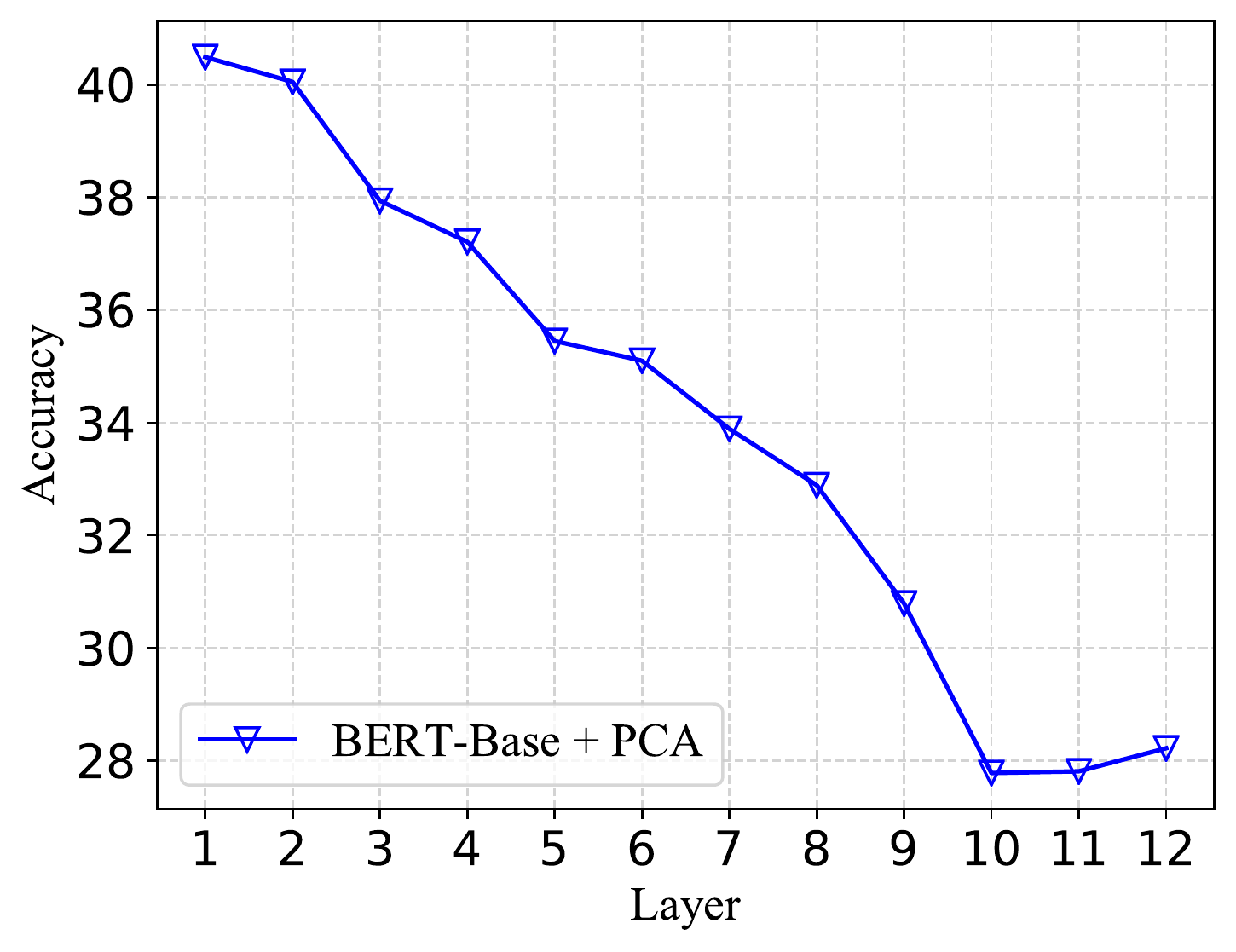}
\caption{Zero-shot accuracy $a\times100$ of using the first principle components from BERT-Base model's different layers as word embeddings on the word analogy task MSR \citep{jastrzebski2017web}. Each layer is evaluated once as no finetuning is required for the pretrained BERT.}
\label{fig:layers_wss}
\end{figure}

%% file: figures/layers/layers.tex
\begin{figure}[!t]
\centering
\includegraphics[width=.84\linewidth]{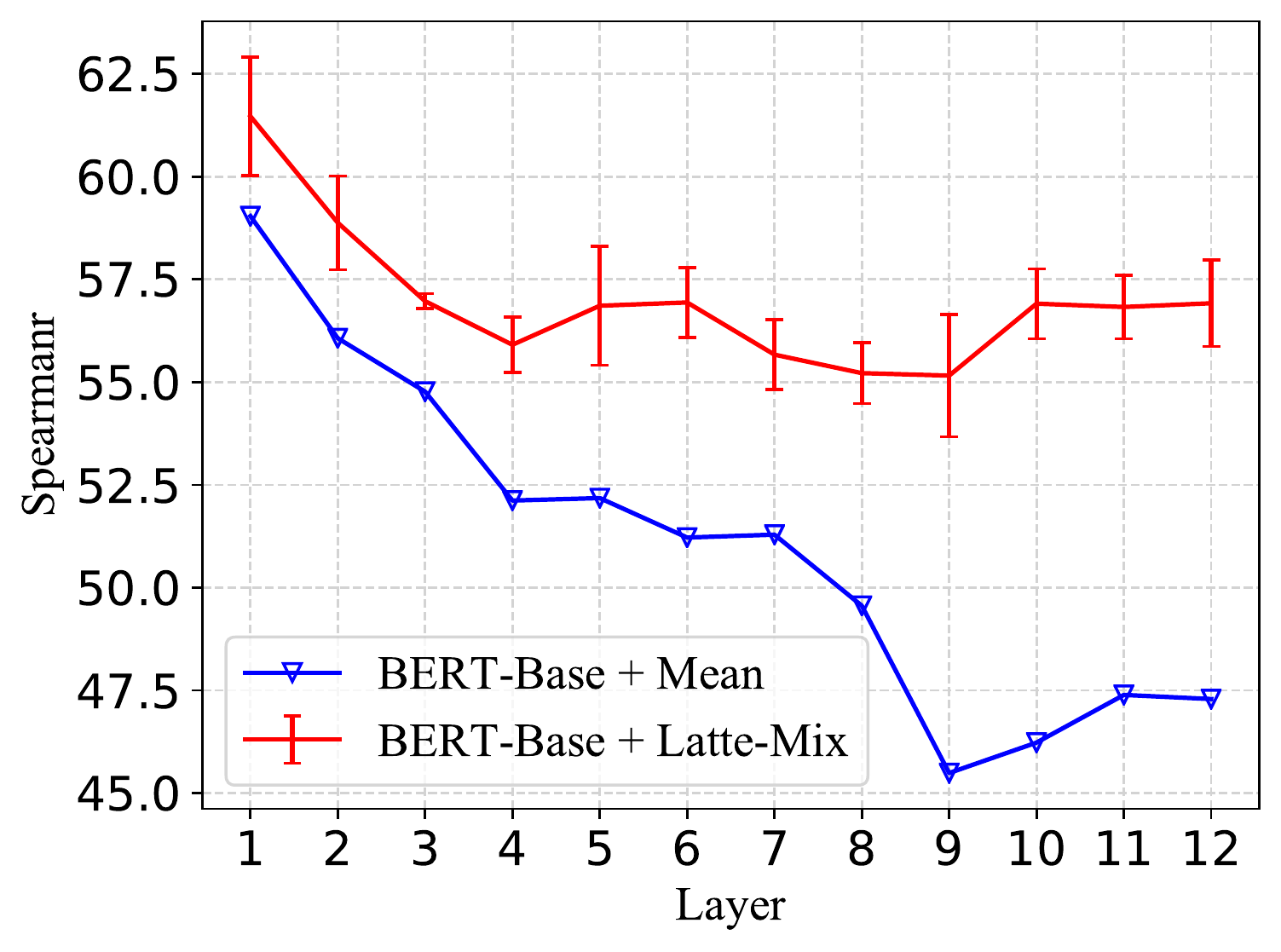}
\caption{Zero-shot spearman's rank correlation $\rho\times100$ between the cosine similarity of BERT-Base model's different layers and the gold labels with different token aggregation strategies on STSbenchmark \citep{cer2017semeval}. Standard deviation of using the VAE for different layers is reported, while mean pooling is a static method which is evaluated once.}
\label{fig:layers_sts}
\end{figure}

%% file: tables/sbert_mean_vs_dist_base.tex
\begin{table}[!b]
  \centering
  {
  \begin{tabular}{l@{\hskip .2in}l@{\hskip 0.15in}l@{\hskip 0.15in}}
\toprule
    Methods &SICK-R &STSb \\\midrule
    SBERT-Base + \texttt{[CLS]} + Cosine   &72.21  &76.19  \\
    SBERT-Base + Max + Cosine  &73.25  &76.74 \\
    SBERT-Base + Mean + Cosine  & 72.91  & 76.98 \\
    SBERT-Base + Latte-Mix + Cosine &74.34 &78.75  \\
    SBERT-Base + Latte-Mix + JS & \underline{74.38} & \underline{79.62}  \\\hline
    SBERT-Large + \texttt{[CLS]} + Cosine  &72.35  &78.15   \\
    SBERT-Large + Max + Cosine  &73.62  &78.30   \\
    SBERT-Large + Mean + Cosine & 73.75  & 79.19  \\
    SBERT-Large + Latte-Mix + Cosine &74.68 &79.82\\ 
    SBERT-Large + Latte-Mix + JS & \textbf{74.86}* & \textbf{80.89}*\\ 
    \bottomrule
  \end{tabular}}
  \caption{Zero-shot spearman's rank correlation $\rho\times100$ between the gold labels and the similarity measurement provided by finetuned models with different token aggregation strategies on SICK-R and STSb datasets. The models are only finetuned on the SNLI dataset without accessing the STS training data. JS: Jason-Shannon Divergence.}
  \label{tbl:sts_sbert_base}
\end{table}

%% file: sections/related.tex
\section{Related Work}
\label{sec:related}
\paragraph{Vector-Based Approaches for Semantics Search} 
Learning low-dimensional vectors from large text corpora has already become the foundation of many modern natural language applications. 
Earliest work such as TF-IDF \citep{gabrilovich2007tfidf} and BM25 \citep{beaulieu1997bm25} consider documents as vectors whose weights of each dimension are computed by the word frequencies. 
It is later popularized by learning algorithms such as Word2Vec \citep{mikolov2013word2vec,pennington2014glove,bojanowski2017fasttext} that embed words to low-dimensional vectors by optimizing unsupervised objectives.
These ideas have quickly been extended to sentences where
sentence embeddings can also be learned from their contexts and other unsupervised tasks \citep{kiros2015skipthought,logeswaran2018quickthought,gan2016convsent,ethayarajh2018usif,le2014sent2vec}. Another type of approaches that leverage pre-trained language models \citep{peters2018elmo,devlin2018bert,liu2019roberta,yang2019xlnet} aggregates the contextual representation of each word in a sentence using mean/max pooling which is rather ineffective. Methods like USE \citep{cer2018use} and Sentence-BERT \citep{reimers2019sbert} further finetune the pre-trained language models on labelled sentence pairs such as the SNLI dataset \citep{bowman2015snli}. More sophisticated token aggregation strategies such as hierarchical pooling \citep{shen2018hierpooling} also yield better results for downstream classification tasks than mean pooling.

Prior to our method, similar Bayesian approaches for developing multi-sense word embedding \citep{barkan2016bayeskipgram,bravzinskas2017bayeskipgram,athiwaratkun2018probfasttext} and word sense disambiguation \citep{gliozzo2004wsd} have been proposed. In comparison, our method does not intend to embed new information but only transforms the original word embedding into a latent space, and the final latent semantics mixtures just provide us a better measurement for the true semantic similarity. In addition, our approach also resembles the Word Mover's Distance \citep{kusner2015wmd} for word semantics alignment but has a lower time complexity compared to solving a discrete transportation problems with constraints.

\paragraph{Variational Autoencoders}
\citet{hinton1994hintonvae} first develop a prototype of modern VAEs, which learns latent factorial codes with minimum discription length. \citet{kingma2013vae} and \citet{rezende2014vae} later propose the reparameterization trick to produce differentiable and low-variance stochastic samples. A common choice for reparameterizing the variational distribution is the Normal distribution, while recently \citet{maddison2016concrete} and \citet{jang2016catvae} propose the Categorical VAE which uses the Gumbel-Softmax distribution to produce categorical samples. 

VAEs have also demonstrated their utilities in many NLP applications such as text generation and representation learning \citep{bowman2015seqvae,li2020optimus}, while one of the biggest challenges is the KL vanishing problem \citep{he2019linearanneal,chen2016lossyvae}. \citet{higgins2016betavae} propose to directly scale the KL loss while \citet{he2019linearanneal} and \citet{fu2019cycanneal} use different schedules for annealing the KL term's coefficient. Other methods such as KL thresholding \citep{kingma2016freenats}, using a conditional prior distribution \citep{goyal2017zforcing}, and inverse auto-regressive flow \citep{chen2016lossyvae,kingma2016iaf} also provide regularization on the latent space of VAEs. 

Additional to VAEs, there are also other types of Bayesian methods for learning the semantics distribution, such as mutual information \citep{kong2019misent} and noise contrastive estimation \citep{oord2018cpc}. Moreover, the implementation of the VAE itself can take in many forms such as using a LSTM encoder \citep{hochreiter1997lstm} or attention-pooling \citep{vaswani2017transformer}. However, such implementation would require using an auto-regressive decoder, which exacerbates the KL vanishing problem empirically \citep{chen2016lossyvae,he2019linearanneal}. In comparison, the token-wise Categorical VAE not only fits nicely in our theoretical framework but also has a simple implementation that works well with pre-trained language models in practice.




%% file: sections/discussion.tex
\section{Conclusion}
\label{sec:conlusion}
In this paper, we propose Latte-Mix, a simple variational approach that learns latent categorical mixtures of word semantics using a VAE model. Our method outperforms conventional token aggregation strategies (e.g., mean pooling) for pre-trained language models on sentence semantics search. We theoretically show that measuring the distance between primitive statistics, such as the mean of word embeddings, can not fully reflect the similarity between sentences' semantics. To solve this problem, we use the distance between the latent categorical mixtures as measurement for sentence semantics, which is more consistent with the true semantic similarity. We also explain under our theoretical framework that finetuning on labelled sentence pairs helps to learn a better distance metric even if the metric itself is theoretically flawed. We evaluate our method on various common benchmarks such as STS, and specifically, the effect of four important variables: the token aggregation method, the word embedding model, the variational distribution, and whether finetuning on labelled sentence pairs. We show that by combining the latest results in these four directions, our method manages to achieve state-of-the-art zero-shot performance on various textual semantics similarity datasets with a minimum number of training steps, as our method only requires learning a small VAE while the pretrained model remains fixed during training.

%% file: sections/appendix.tex
\section{Appendix}
\label{sec:appendix}

\subsection{Proof of Theorem \ref{thm:mean_thm}} \label{thm:mean_proof}
Consider two sentences $s_1$ and $s_2$ with underlying semantics $m_1$ and $m_2$, represented by two sequences of word semantics vectors $\{u^{(i)}_1\}_{i=1}^{K_1}$ and $\{u^{(i)}_2\}_{i=1}^{K_2}$ sampled i.i.d. from two conditional semantics distributions $p(u|m_1)$ and $p(u|m_2)$, respectively. If $m_1 = m_2$, it's obvious that $p(u|m_1) = p(u|m_2)$ and therefore $\mathbb{E}_{p(u\mid m_1)}[u] = \mathbb{E}_{p(u\mid m_2)}[u]$, so we only provide proof for the (2) statement in Theorem \ref{thm:mean_thm}.

\textit{Claim.} If $\mathbb{E}_{p(u\mid m_1)}[u] = \mathbb{E}_{p(u\mid m_2)}[u]$, $m_1$ and $m_2$ are not always equal for some $m_1$ and $m_2$. 

\textit{Proof by Example.} Consider two categorical distributions $p(u\mid m_1)$ and $p(u\mid m_2)$ with the same support $[0, 1, 2]$, where $m_1  \in \mathbb{R}_{+}^3$ and $m_2 \in \mathbb{R}_{+}^3$. Suppose $m_1 = [0, 1, 0]$ and $m_2 = [1, 0, 1]$, and $p$ is a normalization function where $p_i(u\mid m) = \frac{m_{i}}{\sum_j m_{j}}$ for the $i$th class, so we know that $p(u\mid m_1) = [0, 1, 0]$ and $p(u\mid m_2) = [0.5, 0, 0.5]$. Therefore, $\mathbb{E}_{p(u\mid m_1)}[u] = \mathbb{E}_{p(u\mid m_2)}[u] = 1$ but $m_1\neq m_2$, thereby proving that $m_1$ and $m_2$ are not necessarily equal when $\mathbb{E}_{p(u\mid m_1)}[u] = \mathbb{E}_{p(u\mid m_2)}[u]$ for some $m_1$ and $m_2$. Done.

\subsection{Proof of Theorem \ref{thm:dist_thm}} \label{thm:dist_proof}
Consider two sentences $s_1$ and $s_2$ with underlying semantics $m_1$ and $m_2$, represented by two sequences of latent word semantics vectors $\{z^{(i)}_1\}_{i=1}^{K_1}$ and $\{z^{(i)}_2\}_{i=1}^{K_2}$ sampled i.i.d. from two conditional semantics distributions $p(z\mid m_1)$ and $p(z\mid m_2)$, respectively. If $m_1 = m_2$, it's obvious that $p(z\mid m_1) = p(z\mid m_2)$ and therefore $D(p(z\mid m_1) || p(z\mid m_2)) = 0$, so we only provide proof for the other direction.

\textit{Claim.} If $D(p(z\mid m_1) || p(z\mid m_2)) = 0$ and $p_{z\mid M}$ is an invertible mapping from semantics space $M$ to the latent distribution $P$, where $p_{Z\mid M}(m) = p(z|m)$, then $m_1 = m_2$.

\textit{Proof.} Suppose that $D$ is a valid distance metric in the space of $P$, according to the identity of indiscernibles of distance metrics 
, we know that when $D(p(z\mid m_1) || p(z\mid m_2)) = 0$, $p(z\mid m_1) = p(z\mid m_2)$ for every $p(z\mid m_1) \in P$ and $p(z\mid m_2)\in P$. 

If $p_{Z|M}$ is an invertible mapping from $M$ to $P$, there exists its left inverse function $g$ from $P$ to $M$ such that $g(p_{Z|M}(m)) = m$ for every $m \in M$. Therefore, if $p(z\mid m_1) = p(z\mid m_2)$, then $g(p(z\mid m_1)) = g(p(z\mid m_2)) \Rightarrow m_1 = m_2$ for every $m_1 \in M$ and $m_2\in M$. Done.


In addition, one might notice that if $p_{Z\mid M}$ is a Categorical distribution computed by a softmax function of $M$, it will not be an invertible mapping from $M$ to $P$. Indubitably, in this work, the variational distribution $q_{Z\mid M}$ is computed by the softmax function which is not invertible. However, the true posterior $p_{Z\mid M}$ need not to be computed from a softmax function and it is theoretically possible to be generated by any invertible function of $M$ that we might not know of. Therefore, it's valid to assume $p_{Z\mid M}$ is a Categorical distribution without violating Theorem \ref{thm:dist_thm}.

\subsection{Proof of Equation \ref{eq: mi}} \label{thm:mi_proof}
Consider $n$ pairs of sentences $\{(s_1^{(i)}, s_2^{(i)})\}_{i=1}^{n}$ with underlying semantics $\{(m_1^{(i)}, m_2^{(i)})\}_{i=1}^{n}$, and the $i$th sentence pair $(s_1^{(i)}, s_2^{(i)})$ is represented by a pair of mean of the sentences' word embeddings $(\bar u_1^{(i)}, \bar u_2^{(i)})$, where $\bar u^{(i)}\in \mathbb{R}^{c}$ and $c$ is the dimensionality of the mean vectors. Each sentence pair $(s_1^{(i)}, s_2^{(i)})$ is labelled by a categorical human score function $f(m_1^{(i)}, m_2^{(i)})$, where $f:\mathbb{R}^{c}\rightarrow \mathbb{R}^{l}$, $l$ is the number of the score's categories, and $f$ is a one-hot vector.

Suppose a weight matrix $W \in \mathbb{R}^{c\times l}$ with the softmax activation \citep{goodfellow2016deep} is used for the maximum likelihood estimation of $\{f^{(i)}\}_{i=1}^{n}=\{f(m_1^{(i)}, m_2^{(i)})\}_{i=1}^{n}$ given features $\{x^{(i)}\}_{i=1}^{n}$, which could be expressed as 
\begin{linenomath}
\begin{align*}
  \underset{W}{\argmax} \sum_{i=1}^n\ln p(\hat f^{(i)} = f^{(i)}\mid x^{(i)}; W),
\end{align*}
\end{linenomath} 

where $p(\hat f^{(i)}\mid x^{(i)}; W)$ is a Categorical distribution in $\mathbb{R}^l$ and maximizing $\ln p(\hat f^{(i)} = f^{(i)}\mid x^{(i)}; W)$ corresponds to minimizing the cross-entropy loss between the distributions of $f$ and $\hat f$:
\begin{linenomath}
\begin{align}\label{eq: softmax}
  &\underset{W}{\argmax} \sum_{i=1}^n\ln p(\hat f^{(i)} = f^{(i)}\mid x^{(i)}; W)\nonumber \\
  \Rightarrow &\underset{W}{\argmin}-\sum_{i=1}^{n}\sum_{j=1}^{l} f_{j}^{(i)}\ln p(\hat f_{j}^{(i)} = f_{j}^{(i)}\mid x; W),
\end{align}
\end{linenomath}
where $$\hat f^{(i)}_j = \frac{\exp(x^{(i)T}W_j)}{\sum_{j'=1}^n \exp(x^{(i)}TW_{j'})}$$ is the prediction of the $i$th sample for the $j$th category and $W_j$ is the $j$th column of the weight matrix $W$.

As there is not always a closed-form solution for Eq. \eqref{eq: softmax}, a more general method is to use gradient descent \citep{goodfellow2016deep} to approximate the true solution. If so, we have the following statement.

\textit{Claim.} 
If $x = |\bar u_1 - \bar u_2|$, $d(\bar u_1, \bar u_2)$ is the $\ell_2$ distance between mean vectors $\bar u_1$ and $\bar u_2$, and the weight matrix $W$ is initialized with a Normal distribution $\mathcal{N}(0, \sigma^2)$, then 
$$\mathbb{E}[\ln p(\hat f=f\mid d(\bar u_1, \bar u_2))] = \mathbb{E}[\ln p(\hat f=f\mid x; W)].$$

\textit{Proof.} 
The update rule of using gradient descent to solve the optimization problem w.r.t the weight matrix $W$ in Eq. \eqref{eq: softmax} is
\begin{linenomath}\begin{align*}
    W^{(i)}_j = W^{(i-1)}_j - \alpha \ast (\hat f^{(i)}_j - f^{(i)}_j)x^{(i)}, \alpha > 0,
\end{align*}\end{linenomath}
where $W^{(i)}_j$ is the weight vector updated using the $i$th sample $x^{(i)}$ and $\alpha$ is the learning rate. $(\hat f^{(i)}_j - f^{(i)}_j)x^{(i)}$ is the gradient of the objective function in Eq. \eqref{eq: softmax} w.r.t $W_j^{(i-1)}$. The final solution $W^{*}$ after iterating all $n$ samples in the dataset could be expressed as a function of the initial vector $W^{(0)}$:
\begin{linenomath}\begin{align*}
    W^{*} = W^{(0)} + X^T\Delta F_{\alpha},
\end{align*}\end{linenomath}
where $W^{(0)} \sim \mathcal{N}(0, \sigma^2)$, $X \in \mathbb{R}^{n\times c}$, and $\Delta F_{\alpha} \in \mathbb{R}^{n\times l}$. The $i$th row of $X$ is $|\bar u^{(i)}_1 - \bar u^{(i)}_2|^T$ and the $i$th row of $\Delta F_{\alpha}$ is $-\alpha(\hat f^{(i)} - f^{(i)})^T$, so the $j$th column of $X^T\Delta F_{\alpha}$ is $\sum_{i=1}^n -\alpha(\hat f^{(i)}_j - f^{(i)}_j)x^{(i)}$. Therefore, the prediction given a new sample $x \in \mathbb{R}^c$ is
\begin{linenomath}\begin{align*}\label{eq: pred}
    p(\hat f\mid x; W)&= \text{softmax}(x^TW^{*})\nonumber \\
    &= \text{softmax}(x^TW^{(0)} + x^TX^T\Delta F_{\alpha})\nonumber\\
           &= \text{softmax}(x^TW^{(0)} + (Xx)^T\Delta F_{\alpha}).
\end{align*}\end{linenomath}
If $x = |\bar u_1 - \bar u_2|$ and $d$ is the $\ell_2$ distance, then the $i$th row of $Xx$ is 
\begin{linenomath}\begin{align*}
(Xx)^{(i)} = &|\bar u^{(i)}_1 - \bar u^{(i)}_2|^T|\bar u_1 - \bar u_2|\nonumber\\
= &(d(\bar u_1, \bar u_2)^2 + d(\bar u^{(i)}_1, \bar u^{(i)}_2)^2 \nonumber\\&- d(|\bar u^{(i)}_1 - \bar u^{(i)}_2|, |\bar u_1 - \bar u_2|)^2)/2,
\end{align*}\end{linenomath}
which suggests that $(Xx)^T\Delta F_{\alpha}$ is a function of $d(\bar u_1, \bar u_2)$ as well. As $W^{(0)} \sim \mathcal{N}(0, \sigma^2)$, then $x^TW^{(0)}$ could be seen as being sampled from
$$\mathcal{N}(\sum_k x_k\ast 0, \sum_k x_k^2\sigma^2) \Rightarrow \mathcal{N}(0, d(\bar u_1, \bar u_2)^2\sigma^2).$$ Consequently, $x^TW^{(0)}$ could also be expressed as a stochastic function of $d(\bar u_1, \bar u_2)$. Therefore, $p(\hat f = f\mid x; W) = p(\hat f = f\mid d(\bar u_1, \bar u_2))$ and thereby proving $\mathbb{E}[\ln p(\hat f=f\mid d(\bar u_1, \bar u_2))] = \mathbb{E}[\ln p(\hat f=f\mid x; W)]$ if the weight matrix $W$ is initialized with a Normal distribution $\mathcal{N}(0, \sigma^2)$, $x = |\bar u_1 - \bar u_2|$, and $d$ is the $\ell_2$ distance. Done.

\subsection{Model Implementation and Training Details}
For the pretrained language models and word embeddings, we use the existing implementations and pretrained model weights as listed below:
\begin{itemize}
    \item Skip-Gram \citep{mikolov2013word2vec}: Pre-trained vectors trained on part of Google News dataset (about 100 billion tokens). The model contains 300-dimensional vectors for 3 million words and phrases, available at \url{https://code.google.com/archive/p/word2vec/}.
    \item GloVe \citep{pennington2014glove}: Training is performed on the Common Crawl datasets (about 840 million tokens). The model contains 300-dimensional vectors for 2.2 million words and phrases, available at \url{https://github.com/stanfordnlp/GloVe}.
    \item BERT \cite{devlin2018bert} and RoBERTa \citep{liu2019roberta}: Pre-trained language models including BERT-Base, BERT-Large, RoBERTa-Base and RoBERTa-Large, available at \url{https://huggingface.co/transformers/}. 
    \item Sentence-BERT and Sentence-RoBERTa \cite{reimers2019sbert}: Pretrained language models mentioned above which have been further finetuned on the Stanford Natural Language Inference dataset, available at \url{https://github.com/UKPLab/sentence-transformers}.
\end{itemize}
All experiments for the baseline methods are only evaluated once using the pretrained model weights or embeddings from the above sources, and therefore no standard deviation is reported. Other experiments that include a VAE model in this paper are at least run with 3 random seeds and the standard deviations are reported in Tbl. \ref{tbl:hyper_all}, 
\ref{tbl:sts_all},
\ref{tbl:sbert_sts_all},
\ref{tbl:base_layer_sts_all}, \ref{tbl:base_layers_web_all}, \ref{tbl:large_layer_sts_all}, and \ref{tbl:large_layers_web_all}. Training a VAE model using the default hyperparameters mentioned below with the BERT-Base model takes about 1 hour with 1 NVIDIA V100 Tensor Core GPU and 16GB of memory.

To learn the variational disrtribution $q_\phi(z|u)$, we use the final outputs of the pre-trained embeddings as inputs for the VAE as well as the reconstruction target. We freeze the weights of the pre-trained models and train the VAE for 1 epoch using a batch size of 16 and the Adam optimizer \citep{kingma2014adam} with a linear-warmup learning rate scheculed from 2e-5 to 1 and then back to 2e-5. To mitigate the KL vanishing problem, we use the linear-warmup schedule \citep{he2019linearanneal} for the coefficient $\beta$ in Eq. \eqref{eq: elbo} where we gradually anneal $\beta$ from 0 to 1. We also apply the KL thresholding strategy \citep{kingma2016freenats} with a threshold $\epsilon=0.3$ so that the KL loss will not be punished below this threshold. We use a categorical uniform distribution for the latent prior $p_{\theta}(z)$.

\subsection{Experiment Results: Hyperparameters for VAEs}\label{sec:vae_hyper_all}
This section provides additional results for Section \ref{sec:experiments:vae_hyper} as shown in Tbl. \ref{tbl:hyper_all}. 
We perform a hyperparameter search for all VAE models. We compare different choices for the variational distribution for the VAE, including the Normal distribution and Categorical distribution, as well as an autoencoder as the baseline. We further perform hyperparameter search on the latent dimensions, number of latent classes, and scale of the temperature coefficient for the Categorical VAE. In Section \ref{sec:experiments:vae_hyper} we have explained that using an 1-layer encoder for the VAE could mitigate the KL vanishing problem. We also find that it's important to use an expressive decoder (in our case 3 layers) to minimize the reconstruction error. For the categorical VAE, we find that using a large number of latent classes such as 100 is more beneficial, as each mode in the latent categorical mixture has a higher chance to be separated from each other and therefore it is easier to align different mixture distributions. Similar tuning for highlighting the modes in the mixture distribution has been mentioned in Section \ref{sec:experiments:vae_hyper}, where we use a small temperature in Eq. \ref{eq: gumbel} such as 0.3 for training and test. Another important factor that affects the results in Tbl. \ref{tbl:base_layer_sts_all} is the metric for measuring the distance between two distributions/embeddings. We could see that for the Latte-Mix, the cosine distance outperforms Jason-Shannon divergence and $\ell_2$ distance on most semantics similarity tasks and in most hyperparameter settings. For the Normal VAE, we only report the results of using the $\ell_2$ distance as there's no closed form solution for Jason-Shannon divergence between two Gaussian mixtures and Monte-Carlo yields poor approximation in high dimensional space. 

\subsection{Experiment Results: STS} \label{sec:sts_all}
This section provides additional results for Section \ref{sec:experiments:main} as shown in Tbl. \ref{tbl:sts_all}. We mainly compare the performance of our method with other token-aggregation approaches based on the pretrained language models such as BERT and RoBERTa as well as static embeddings such as Glove and Skip-Gram in Tbl. \ref{tbl:sts_all}. 

We also train the VAE model with the SNLI-finetuned Sentence-BERT and Sentence-RoBERTa model \citep{reimers2019sbert}. The results are reported in Tbl. \ref{tbl:sbert_sts_all}. As the authors of Sentence-BERT have not released their evaluation pipeline on some of the STS tasks such as STS2012-2016, we use the data and pipeline from the uSIF \citep{ethayarajh2018usif} paper's repository: \url{https://github.com/kawine/usif}. Our results are a little bit different from \citep{reimers2019sbert} but consistent on most of the tasks.

We also report the evaluation results using different metrics for measuring the distance between the latent mixture distributions from the VAEs, including the $\ell_2$ distance and Jason-Shannon Divergence. For the Categorical VAE, we also measure the distributional distance using the cosine similarity, as the categorical distributions are represented by finite dimensional vectors.

\subsection{Experiment Results: WEB} \label{sec:web_all}
This section provides additional results for Section \ref{sec:experiments:bert_layers} as shown in Tbl. \ref{tbl:base_layer_sts_all}, \ref{tbl:base_layers_web_all}, \ref{tbl:large_layer_sts_all}, and \ref{tbl:large_layers_web_all}. We take the first principle component of BERT-Base's and BERT-Large's different layers as word embeddings and evaluate their qualities on the word embedding benchmark (WEB) \citep{jastrzebski2017web}. The principle component is computed using singular value decomposition on the STS data. We use the evaluation pipeline provided in this repo: \url{https://github.com/kudkudak/word-embeddings-benchmarks/tree/master/web}. The results are reported in Tbl. \ref{tbl:base_layers_web_all} and \ref{tbl:large_layers_web_all}. We also use these pretrained models' different layers as word embeddings and evaluate their combination with different token aggregation methods on the STS datasets. The results are reported in Tbl. \ref{tbl:base_layer_sts_all} and \ref{tbl:large_layer_sts_all}. The quality of the word embeddings shows a positive correlation with the performance of the corresponding model on the semantics textual similarity datasets.

\input{tables/vae_hyper_all}

\input{tables/mean_vs_dist_all}
\input{tables/sbert_mean_vs_dist_all}

\input{tables/bert_base_layers_sts_all}
\input{tables/bert_base_layers_web_all}
\input{tables/bert_large_layers_sts_all}

\input{tables/bert_large_layers_web_all}

%% file: tables/vae_hyper_all.tex
\begin{table*}
  \centering
\resizebox{17cm}{!}{
  \begin{tabular}{l@{\hskip .15in}l@{\hskip 0.15in}l@{\hskip 0.15in}l@{\hskip 0.15in}l@{\hskip 0.15in}l@{\hskip 0.15in}l@{\hskip 0.15in}l@{\hskip 0.15in}}
\toprule

Methods &STS12 &STS13 &STS14 &STS15 &STS16 &SICK-R &STSb \\\midrule
AE (1-layer) + Cosine  & 26.37$\pm$0.52 & \textbf{55.81$\pm$1.34} & \textbf{37.21$\pm$0.92} & 50.26$\pm$0.45 & \textbf{52.21$\pm$0.48} & \textbf{53.9$\pm$0.51}  & \textbf{37.18$\pm$0.7}  \\
AE (2-layer) + Cosine                         & 27.08$\pm$2.14 & 53.81$\pm$2.24 & 35.99$\pm$0.18 & \textbf{50.79$\pm$1.0}  & 50.26$\pm$0.43 & 53.82$\pm$0.93 & 36.89$\pm$1.18 \\
AE (3-layer) + Cosine                         & \textbf{28.31$\pm$3.33} & 54.6$\pm$0.71  & 35.32$\pm$0.65 & 49.52$\pm$0.51 & 49.77$\pm$0.86 & 52.93$\pm$0.65 & 36.79$\pm$0.3  \\\hline
Normal-VAE (1-layer) + L2       & \textbf{42.71$\pm$1.33} & 56.5$\pm$0.74  & \textbf{47.19$\pm$1.73} & \textbf{62.22$\pm$1.51} & \textbf{60.94$\pm$0.65} & \textbf{55.91$\pm$1.14} & \textbf{50.84$\pm$2.09} \\
Normal-VAE (2-layer) + L2       & 31.21$\pm$5.46 & 58.76$\pm$0.74 & 43.7$\pm$1.43  & 55.16$\pm$0.39 & 57.19$\pm$3.42 & 54.85$\pm$1.43 & 44.3$\pm$0.58  \\
Normal-VAE (3-layer) + L2       & 28.97$\pm$2.72 & \textbf{60.85$\pm$1.78} & 41.45$\pm$0.79 & 52.48$\pm$0.7  & 55.22$\pm$0.81 & 53.76$\pm$0.43 & 40.3$\pm$1.67  \\\hline
Latte-Mix (1-layer) + Cosine                       & \textbf{42.7$\pm$0.74}  & 66.54$\pm$0.41 & \textbf{52.67$\pm$0.62} & \textbf{68.56$\pm$0.4}  & \textbf{63.84$\pm$0.29} & \textbf{60.94$\pm$0.57} & \textbf{56.6$\pm$1.02}  \\
Latte-Mix (1-layer) + JS                       & 36.25$\pm$0.3  & \textbf{67.67$\pm$0.27} & 51.25$\pm$0.19 & 66.37$\pm$0.12 & 63.39$\pm$0.25 & 59.89$\pm$0.31 & 52.61$\pm$0.08 \\
Latte-Mix (1-layer) + L2                       & 27.43$\pm$1.44 & 67.15$\pm$0.54 & 44.0$\pm$0.77  & 45.66$\pm$1.6  & 51.88$\pm$0.26 & 53.68$\pm$0.09 & 42.6$\pm$0.81  \\\hline
Latte-Mix (2-layer) + Cosine                       & \textbf{32.56$\pm$0.53} & 64.17$\pm$0.63 & 45.5$\pm$0.17  & 58.56$\pm$0.8  & 57.67$\pm$0.82 & 55.96$\pm$1.23 & 45.22$\pm$1.03 \\
Latte-Mix (2-layer) + JS                       & 32.59$\pm$0.6  & \textbf{64.67$\pm$0.82} & \textbf{45.9$\pm$0.23}  & \textbf{58.82$\pm$0.94} & \textbf{57.75$\pm$1.15} & \textbf{56.23$\pm$0.96} & \textbf{45.39$\pm$0.71} \\
Latte-Mix (2-layer) + L2                       & 32.06$\pm$0.3  & 63.85$\pm$1.11 & 44.74$\pm$0.28 & 57.69$\pm$0.91 & 55.72$\pm$1.08 & 55.72$\pm$1.21 & 43.48$\pm$1.08 \\\hline
Latte-Mix (3-layer) + Cosine                       & \textbf{26.9$\pm$2.19}  & \textbf{66.16$\pm$0.39} & \textbf{43.89$\pm$1.35} & \textbf{52.48$\pm$1.27} & \textbf{58.43$\pm$1.07} & \textbf{55.86$\pm$0.89} & \textbf{41.49$\pm$0.99} \\
Latte-Mix (3-layer) + JS                       & 25.55$\pm$2.88 & 66.06$\pm$0.27 & 42.8$\pm$1.65  & 50.95$\pm$1.66 & 58.07$\pm$1.1  & 55.8$\pm$1.18  & 40.34$\pm$1.02 \\
Latte-Mix (3-layer) + L2                       & 22.09$\pm$3.67 & 64.54$\pm$0.77 & 39.42$\pm$1.73 & 43.93$\pm$2.75 & 54.24$\pm$2.01 & 54.15$\pm$1.14 & 35.43$\pm$2.04 \\\hline
Latte-Mix (20-class) + Cosine               & \textbf{41.84$\pm$0.46} & 67.43$\pm$0.63 & \textbf{52.72$\pm$0.43} & \textbf{67.43$\pm$0.42} & \textbf{63.57$\pm$0.21} & \textbf{60.43$\pm$0.34} & \textbf{55.67$\pm$1.04} \\
Latte-Mix (20-class) + JS               & 36.2$\pm$0.32  & \textbf{69.12$\pm$0.23} & 51.23$\pm$0.24 & 63.88$\pm$0.36 & 62.48$\pm$0.29 & 59.36$\pm$0.23 & 52.45$\pm$0.02 \\
Latte-Mix (20-class) + L2               & 33.31$\pm$0.87 & 68.36$\pm$0.65 & 48.27$\pm$0.46 & 55.66$\pm$1.03 & 58.02$\pm$0.38 & 56.33$\pm$0.48 & 48.34$\pm$0.19 \\\hline
Latte-Mix (50-class) + Cosine               & \textbf{41.95$\pm$0.42} & 67.49$\pm$0.76 & \textbf{53.21$\pm$0.33} & \textbf{68.03$\pm$0.26} & \textbf{63.4$\pm$0.1}   & \textbf{61.17$\pm$0.24} & \textbf{56.51$\pm$0.24} \\
Latte-Mix (50-class) + JS               & 36.2$\pm$0.32  & \textbf{68.69$\pm$0.31} & 51.69$\pm$0.22 & 65.66$\pm$0.33 & 63.02$\pm$0.3  & 60.18$\pm$0.3  & 52.65$\pm$0.24 \\
Latte-Mix (50-class) + L2               & 29.59$\pm$0.31 & 67.95$\pm$0.31 & 46.25$\pm$0.07 & 49.34$\pm$0.67 & 54.4$\pm$0.21  & 55.27$\pm$0.35 & 45.33$\pm$0.43 \\\hline
Latte-Mix (100-class) + Cosine             & \textbf{42.82$\pm$0.2}  & 66.65$\pm$0.2  & \textbf{53.0$\pm$0.25}  & \textbf{68.63$\pm$0.23} & \textbf{63.69$\pm$0.29} & \textbf{61.37$\pm$0.12} & \textbf{56.98$\pm$0.34} \\
Latte-Mix (100-class) + JS            & 36.05$\pm$0.21 & \textbf{67.91$\pm$0.05} & 51.37$\pm$0.19 & 66.32$\pm$0.22 & 63.42$\pm$0.07 & 60.12$\pm$0.11 & 52.52$\pm$0.19 \\
Latte-Mix (100-class) + L2             & 26.23$\pm$0.31 & 67.29$\pm$0.28 & 43.2$\pm$0.17  & 44.02$\pm$0.53 & 51.7$\pm$0.48  & 53.58$\pm$0.36 & 41.42$\pm$0.25 \\\hline
Latte-Mix (200-class) + Cosine             & \textbf{42.45$\pm$0.92} & \textbf{66.08$\pm$0.7}  & \textbf{52.7$\pm$0.43}  & \textbf{68.97$\pm$0.52} & \textbf{63.57$\pm$0.66} & \textbf{61.01$\pm$0.6}  & \textbf{56.39$\pm$1.5}  \\
Latte-Mix (200-class) + JS             & 36.24$\pm$0.58 & 66.96$\pm$0.13 & 51.12$\pm$0.13 & 67.09$\pm$0.16 & 63.42$\pm$0.35 & 59.92$\pm$0.22 & 52.3$\pm$0.14  \\
Latte-Mix (200-class) + L2             & 25.41$\pm$1.52 & 66.26$\pm$0.11 & 42.14$\pm$0.91 & 41.92$\pm$1.6  & 48.75$\pm$0.9  & 52.31$\pm$0.18 & 39.5$\pm$1.35  \\\hline
Latte-Mix (16-dim) + Cosine                   & \textbf{42.47$\pm$0.93} & \textbf{66.01$\pm$1.13} & \textbf{52.68$\pm$0.81} & \textbf{67.77$\pm$0.57} & \textbf{62.66$\pm$0.88} & \textbf{60.59$\pm$0.3}  & \textbf{56.28$\pm$0.9}  \\
Latte-Mix (16-dim) + JS                   & 35.73$\pm$0.73 & 67.64$\pm$0.61 & 51.36$\pm$0.31 & 65.95$\pm$0.4  & 63.15$\pm$0.42 & 59.79$\pm$0.2  & 52.39$\pm$0.52 \\
Latte-Mix (16-dim) + L2                   & 26.09$\pm$0.57 & 66.57$\pm$0.4  & 43.5$\pm$0.76  & 43.68$\pm$0.96 & 51.23$\pm$0.56 & 52.73$\pm$0.32 & 40.73$\pm$0.85 \\\hline
Latte-Mix (32-dim) + Cosine                   & \textbf{43.04$\pm$0.35} & 67.0$\pm$0.59  & \textbf{53.34$\pm$0.23} & \textbf{68.64$\pm$0.49} & \textbf{63.31$\pm$0.11} & \textbf{61.28$\pm$0.35} & \textbf{56.48$\pm$0.6}  \\
Latte-Mix (32-dim) + JS                   & 36.21$\pm$0.12 & \textbf{67.93$\pm$0.19} & 51.56$\pm$0.08 & 66.2$\pm$0.4   & 63.08$\pm$0.21 & 60.07$\pm$0.22 & 52.26$\pm$0.49 \\
Latte-Mix (32-dim) + L2                   & 26.54$\pm$0.19 & 66.84$\pm$0.06 & 43.75$\pm$0.44 & 44.22$\pm$0.74 & 51.35$\pm$0.2  & 53.46$\pm$0.28 & 41.4$\pm$0.36  \\\hline
Latte-Mix (64-dim) + Cosine                   & \textbf{42.86$\pm$0.5}  & 66.99$\pm$0.36 & \textbf{53.56$\pm$0.25} & \textbf{69.26$\pm$0.15} & \textbf{63.78$\pm$0.29} & \textbf{61.53$\pm$0.12} & \textbf{57.41$\pm$0.11} \\
Latte-Mix (64-dim) + JS                   & 36.19$\pm$0.15 & \textbf{67.95$\pm$0.1}  & 51.57$\pm$0.08 & 66.68$\pm$0.24 & 63.42$\pm$0.1  & 60.23$\pm$0.01 & 52.78$\pm$0.26 \\
Latte-Mix (64-dim) + L2                   & 26.79$\pm$0.21 & 67.38$\pm$0.16 & 44.09$\pm$0.37 & 44.89$\pm$0.23 & 51.76$\pm$0.17 & 53.75$\pm$0.13 & 42.22$\pm$0.57 \\\hline
Latte-Mix (128-dim) + Cosine                & \textbf{43.24$\pm$0.14} & 67.36$\pm$0.15 & \textbf{53.64$\pm$0.13} & \textbf{69.36$\pm$0.26} & \textbf{63.93$\pm$0.04} & \textbf{61.67$\pm$0.18} & \textbf{57.25$\pm$0.31} \\
Latte-Mix (128-dim) + JS                 & 36.39$\pm$0.02 & \textbf{68.02$\pm$0.07} & 51.7$\pm$0.11  & 66.84$\pm$0.21 & 63.41$\pm$0.13 & 60.22$\pm$0.12 & 52.77$\pm$0.19 \\
Latte-Mix (128-dim) + L2                 & 27.28$\pm$0.2  & 67.61$\pm$0.04 & 44.41$\pm$0.12 & 45.47$\pm$0.48 & 51.87$\pm$0.62 & 54.01$\pm$0.18 & 42.25$\pm$0.38 \\\hline
Latte-Mix (0.03-tau) + Cosine & \textbf{42.35$\pm$0.84} & 67.1$\pm$0.27  & \textbf{52.77$\pm$0.08} & \textbf{68.88$\pm$0.46} & \textbf{63.34$\pm$0.3}  & \textbf{61.34$\pm$0.2}  & \textbf{56.64$\pm$0.53}\\
Latte-Mix (0.03-tau) + JS & 36.04$\pm$0.66 & \textbf{68.01$\pm$0.05} & 51.29$\pm$0.09 & 66.49$\pm$0.14 & 63.04$\pm$0.17 & 60.06$\pm$0.1  & 52.74$\pm$0.51\\
Latte-Mix (0.03-tau) + L2 & 27.83$\pm$1.06 & 67.16$\pm$0.32 & 44.32$\pm$0.54 & 47.01$\pm$1.27 & 51.68$\pm$0.84 & 53.67$\pm$0.24 & 42.71$\pm$0.6 \\\hline
Latte-Mix (0.3-tau) + Cosine & \textbf{41.11$\pm$0.5}  & 66.03$\pm$0.39 & \textbf{52.75$\pm$0.43} & \textbf{68.81$\pm$0.39} & \textbf{63.49$\pm$0.43} & \textbf{60.72$\pm$0.19} & \textbf{55.6$\pm$0.51}  \\
Latte-Mix (0.3-tau) + JS & 36.37$\pm$0.41 & \textbf{67.63$\pm$0.16} & 51.55$\pm$0.13 & 67.05$\pm$0.38 & 63.01$\pm$0.4  & 60.04$\pm$0.08 & 52.33$\pm$0.3  \\
Latte-Mix (0.3-tau) + L2 & 27.95$\pm$0.46 & 67.01$\pm$0.13 & 44.56$\pm$0.3  & 44.76$\pm$0.59 & 51.74$\pm$0.54 & 53.78$\pm$0.08 & 41.98$\pm$0.53 \\\hline
Latte-Mix (1-tau) + Cosine       & \textbf{31.35$\pm$1.71} & 67.09$\pm$0.47 & 47.33$\pm$0.28 & 60.18$\pm$0.92 & \textbf{61.52$\pm$0.26} & 56.41$\pm$0.6  & 47.31$\pm$0.82 \\
Latte-Mix (1-tau) + JS       & 31.17$\pm$0.37 & \textbf{68.14$\pm$0.62} & \textbf{48.28$\pm$0.42} & \textbf{61.68$\pm$0.58} & 61.33$\pm$0.32 & \textbf{57.85$\pm$0.14} & \textbf{48.28$\pm$0.21} \\
Latte-Mix (1-tau) + L2       & 31.49$\pm$0.59 & 67.75$\pm$0.18 & 46.4$\pm$0.3   & 53.73$\pm$0.3  & 54.36$\pm$0.78 & 54.75$\pm$0.27 & 44.97$\pm$0.65 \\\hline
Latte-Mix (2-tau) + Cosine       & 27.82$\pm$0.73 & 67.68$\pm$0.53 & 47.32$\pm$1.74 & 58.01$\pm$1.64 & 60.3$\pm$0.85  & 56.06$\pm$0.23 & 44.66$\pm$0.98 \\
Latte-Mix (2-tau) + JS       & 30.19$\pm$0.35 & \textbf{68.02$\pm$0.12} & \textbf{48.39$\pm$0.61} & \textbf{61.06$\pm$0.75} & \textbf{61.03$\pm$0.19} & \textbf{58.0$\pm$0.16}  & \textbf{47.21$\pm$0.47} \\
Latte-Mix (2-tau) + L2       & \textbf{31.53$\pm$0.47} & 67.54$\pm$0.52 & 46.99$\pm$0.37 & 54.23$\pm$0.19 & 54.49$\pm$0.24 & 55.12$\pm$0.04 & 45.03$\pm$0.37 \\
\bottomrule
\end{tabular}
}
  \caption{Zero-shot spearman's rank correlation $\rho\times 100$ of the negative distance between sentence embeddings/distributions and the gold labels on STS tasks. Results of using the BERT-Base model and different hyperparameters such the latent dimension, latent class, temperature coefficient, encoder layers, and distance metrics are reported. AE: Autoencoder. Normal-VAE: Normal VAE. Latte-Mix: Categorical VAE. Cosine: Cosine similarity. L2: $\ell_2$ distance. JS: Jason-Shannon Divergence. N-layer: number of VAE's encoder layers. N-class: number of latent classes. N-dim: number of latent dimensions. N-tau: scale of the temperature coefficient.}
  \label{tbl:hyper_all}
\end{table*}

%% file: tables/mean_vs_dist_all.tex
\begin{table*}
  \centering
\resizebox{18cm}{!}{
  \begin{tabular}{l@{\hskip .15in}l@{\hskip 0.15in}l@{\hskip 0.15in}l@{\hskip 0.15in}l@{\hskip 0.15in}l@{\hskip 0.15in}l@{\hskip 0.15in}l@{\hskip 0.15in}}
\toprule

Methods &STS12 &STS13 &STS14 &STS15 &STS16 &SICK-R &STSb \\\midrule
GloVe + Max + Cosine &48.21  &65.45  &54.63 &67.77  &57.27  &51.48   &58.32   \\
GloVe + Mean + Cosine                & 53.75  & 65.74  & 60.23  & 72.45  & 66.74  & 55.38  & 61.54  \\
GloVe + Latte-Mix + Cosine           & \textbf{55.24$\pm$0.46} & 61.5$\pm$0.82  & \textbf{65.37$\pm$0.23} & \textbf{75.97$\pm$0.18} & 68.35$\pm$0.19 & \textbf{56.49$\pm$0.28} & \textbf{65.84$\pm$0.25} \\
GloVe + Latte-Mix + JS           & 53.79$\pm$0.2  & \textbf{66.88$\pm$0.23} & 65.17$\pm$0.09 & 74.55$\pm$0.07 & 66.46$\pm$0.06 & 55.49$\pm$0.1  & 64.13$\pm$0.1  \\
GloVe + Latte-Mix + L2           & 46.62$\pm$0.21 & \textbf{68.39$\pm$0.18} & 58.93$\pm$0.02 & 65.82$\pm$0.26 & 58.22$\pm$0.22 & 49.45$\pm$0.13 & 58.35$\pm$0.19 \\\hline
Skip-Gram + Max + Cosine &40.26  &61.44  &51.28  &65.80  &56.24  &54.85   &54.45   \\
Skip-Gram + Mean + Cosine             & 46.35   & 60.63  & 59.03  & 73.49  & 60.77  & 56.21  & 57.81  \\
Skip-Gram + Latte-Mix + Cosine        & \textbf{46.5$\pm$0.33} & 62.12$\pm$0.58 & \textbf{60.99$\pm$0.16} & \textbf{75.18$\pm$0.35} & \textbf{63.93$\pm$0.36} & \textbf{59.06$\pm$0.47} & \textbf{58.02$\pm$0.16}  \\
Skip-Gram + Latte-Mix + JS        & 42.58$\pm$0.09 & \textbf{64.58$\pm$0.09} & 55.91$\pm$0.03 & 69.55$\pm$0.04 & 57.31$\pm$0.26 & 55.18$\pm$0.21 & 53.89$\pm$0.04 \\
Skip-Gram + Latte-Mix + L2        & 38.81$\pm$0.34 & 63.38$\pm$0.11 & 49.92$\pm$0.39 & 61.8$\pm$0.57  & 52.01$\pm$0.21 & 51.2$\pm$0.53  & 50.31$\pm$0.22 \\\hline
BERT-Base + \texttt{[CLS]} + Cosine          &21.53  &46.48  &21.27  &37.88  &44.25  &42.42   &20.29   \\
BERT-Base + Max + Cosine          &46.26  &60.02  &46.62  &60.02  &57.78  &57.98   &50.94   \\
BERT-Base + Mean + Cosine          & 30.88  & 65.78  & 47.74  & 60.29  & 63.73  & 58.22  & 47.29  \\
BERT-Base + Latte-Mix + Cosine      & \textbf{43.05$\pm$0.35} & 66.91$\pm$0.24 & \textbf{53.31$\pm$0.38} & \textbf{68.71$\pm$0.13} & 62.88$\pm$0.5  & \textbf{61.26$\pm$0.15} & \textbf{56.82$\pm$0.25} \\
BERT-Base + Latte-Mix + JS      & 36.03$\pm$0.32 & \textbf{67.86$\pm$0.2}  & 51.45$\pm$0.07 & 66.42$\pm$0.07 & \textbf{63.15$\pm$0.25} & 60.08$\pm$0.02 & 52.44$\pm$0.01 \\
BERT-Base + Latte-Mix + L2 & 26.85$\pm$0.49 & 67.37$\pm$0.1  & 43.72$\pm$0.2  & 44.24$\pm$0.47 & 51.85$\pm$0.42 & 53.46$\pm$0.13 & 41.66$\pm$0.32 \\\hline
BERT-Large + \texttt{[CLS]} + Cosine         &27.74  &36.32  &22.59  &29.98  &42.75  &43.43   &26.75   \\ 
BERT-Large + Max + Cosine          &33.76  &62.24  &40.12  &45.07  &57.15  &51.91   &47.18   \\
BERT-Large + Mean + Cosine          & 27.69  & 63.97  & 44.48  & 51.67  & 61.85  & 53.85  & 47.0   \\
BERT-Large + Latte-Mix + Cosine     & \textbf{39.99$\pm$0.3}  & 66.58$\pm$0.25 & \textbf{52.55$\pm$0.25} & \textbf{61.4$\pm$0.2}   & \textbf{65.38$\pm$0.72} & \textbf{56.71$\pm$0.25} & \textbf{56.37$\pm$0.73} \\
BERT-Large + Latte-Mix + JS & 35.61$\pm$0.17 & \textbf{69.53$\pm$0.16} & 50.77$\pm$0.08 & 57.38$\pm$0.09 & 63.13$\pm$0.36 & 55.08$\pm$0.15 & 53.08$\pm$0.29 \\
BERT-Large + Latte-Mix + L2 & 26.5$\pm$0.21  & 66.62$\pm$0.15 & 42.4$\pm$0.15  & 37.81$\pm$0.57 & 53.55$\pm$0.58 & 46.73$\pm$0.25 & 42.84$\pm$0.29 \\\hline
RoBERTa-Base + \texttt{[CLS]} + Cosine          &14.90  &15.08  &28.07  &42.58  &50.43  &63.56   &45.41   \\
RoBERTa-Base + Max + Cosine          &23.84  &54.57  &41.58  &56.51  &55.94  &59.70   &50.01   \\
RoBERTa-Base + Mean + Cosine        & 30.61  & 57.25  & 46.83  & 62.51  & 61.37  & 62.22  & 54.53  \\
RoBERTa-Base + Latte-Mix + Cosine   & \textbf{33.91$\pm$1.17} & \textbf{62.97$\pm$0.51} & \textbf{52.17$\pm$0.49} & 66.52$\pm$0.7  & 62.84$\pm$0.47 & 63.88$\pm$0.56 & \textbf{58.67$\pm$0.87} \\
RoBERTa-Base + Latte-Mix + JS   & 31.94$\pm$0.18 & 61.46$\pm$0.33 & 50.53$\pm$0.17 & \textbf{66.7$\pm$0.23}  & \textbf{62.99$\pm$0.3}  & \textbf{63.96$\pm$0.11} & 57.57$\pm$0.39 \\
RoBERTa-Base + Latte-Mix + L2 & 24.62$\pm$0.33 & 64.69$\pm$0.68 & 47.76$\pm$0.3  & 60.91$\pm$0.2  & 61.33$\pm$0.77 & 62.69$\pm$0.38 & 52.69$\pm$0.42 \\\hline
RoBERTa-Large + \texttt{[CLS]} + Cosine          &0.32  &2.74  &7.83  &22.16  &22.12  &39.88   &12.52   \\
RoBERTa-Large + Max + Cosine          &16.99  &26.42  &28.40  &49.86  &45.59  &45.85   &28.41   \\
RoBERTa-Large + Mean + Cosine       & 26.0   & 50.71  & 44.06  & 62.2   & 60.43  & 58.15  & 47.07  \\
RoBERTa-Large + Latte-Mix + Cosine  & \textbf{28.45$\pm$0.77} & \textbf{57.3$\pm$0.45}  & \textbf{49.68$\pm$0.33} & \textbf{65.96$\pm$0.29} & 65.06$\pm$0.51 & \textbf{59.42$\pm$0.29} & \textbf{53.91$\pm$0.49} \\
RoBERTa-Large + Latte-Mix + JS  & 24.82$\pm$0.11 & 56.58$\pm$0.43 & 45.9$\pm$0.19  & 63.55$\pm$0.17 & 62.69$\pm$0.33 & 58.1$\pm$0.12  & 49.06$\pm$0.29 \\
RoBERTa-Large + Latte-Mix + L2 & 22.29$\pm$0.23 & 64.08$\pm$0.36 & 45.06$\pm$0.21 & 60.24$\pm$0.34 & 62.23$\pm$0.59 & 59.35$\pm$0.33 & 49.53$\pm$0.21 \\ \bottomrule
  \end{tabular}
}
  \caption{Zero-shot spearman's rank correlation $\rho\times 100$ of the negative distance between sentence embeddings/distributions and the gold labels on STS tasks. For BERT and RoBERTa, outputs at the last layer are used as inputs for the token aggregation methods. STS12-STS16: SemEval 2012-2016, STSb: STSbenchmark, SICK-R: SICK relatedness dataset. Cosine: Cosine similarity. Mean: Mean pooling. Max: Max pooling. \texttt{[CLS]}: the embedding for the \texttt{[CLS]} token. Latte-Mix: Latent mixtures from the Categorical VAE model. Cosine: Cosine similarity. L2: $\ell_2$ distance. JS: Jason-Shannon Divergence.}
  \label{tbl:sts_all}
\end{table*}

%% file: tables/sbert_mean_vs_dist_all.tex
\begin{table*}
  \centering
\resizebox{18cm}{!}{
  \begin{tabular}{l@{\hskip .15in}l@{\hskip 0.15in}l@{\hskip 0.15in}l@{\hskip 0.15in}l@{\hskip 0.15in}l@{\hskip 0.15in}l@{\hskip 0.15in}l@{\hskip 0.15in}}
\toprule

Methods &STS12 &STS13 &STS14 &STS15 &STS16 &SICK-R &STSb \\\midrule
SBERT-Base + \texttt{[CLS]} + Cosine          &69.40  &76.73  &72.45  &77.79  &73.21  &72.21   &76.02   \\
SBERT-Base + Max + Cosine          &72.51  &79.78  &73.85  &79.47  &74.45  &73.25   &76.73   \\
SBERT-Base + Mean + Cosine         & 70.97  & 77.49  & 73.19  & 79.09  & 74.27  & 72.91  & 76.98  \\
SBERT-Base+ Latte-Mix + Cosine     & 69.8$\pm$0.77  & 78.42$\pm$0.16 & 75.79$\pm$0.34 & 81.54$\pm$0.39 & 75.65$\pm$0.56 & 74.34$\pm$0.37 & 78.75$\pm$0.38 \\
SBERT-Base + Latte-Mix + JS     & \textbf{70.29$\pm$0.34} & \textbf{80.56$\pm$0.14} & \textbf{76.93$\pm$0.17} & \textbf{82.76$\pm$0.28} & \textbf{76.93$\pm$0.12} & \textbf{74.38$\pm$0.55} & \textbf{79.62$\pm$0.14} \\
SBERT-Base + Latte-Mix + L2     & 67.31$\pm$0.74 & 78.69$\pm$0.23 & 71.94$\pm$0.84 & 67.58$\pm$2.13 & 68.54$\pm$0.43 & 65.43$\pm$1.18 & 72.32$\pm$0.73 \\\hline
SBERT-Large + \texttt{[CLS]} + Cosine          &72.34  &78.52  &74.26  &80.22  &74.82  &72.35   &78.15   \\
SBERT-Large + Max + Cosine          &\textbf{72.66}  &80.40  &75.23  &81.39  &75.37  &73.62   &78.30   \\
SBERT-Large + Mean + Cosine         & 72.27  & 79.1   & 74.9   & 80.99  & 76.24  & 73.75  & 79.19  \\
SBERT-Large+ Latte-Mix + Cosine    & 71.73$\pm$0.31 & 79.25$\pm$0.28 & 76.857 & 82.55$\pm$0.19 & 76.94$\pm$0.48 & 74.68$\pm$0.38 & 79.82$\pm$0.5  \\
SBERT-Large + Latte-Mix + JS    & 72.51$\pm$0.05 & \textbf{81.28$\pm$0.29} & \textbf{77.99$\pm$0.05} & \textbf{83.89$\pm$0.14} & \textbf{78.13$\pm$0.19} & \textbf{74.86$\pm$0.8}  & \textbf{80.89$\pm$0.23} \\
SBERT-Large + Latte-Mix + L2    & 70.21$\pm$1.22 & 76.11$\pm$0.49 & 71.17$\pm$1.09 & 68.75$\pm$3.02 & 70.53$\pm$0.69 & 68.18$\pm$2.04 & 73.16$\pm$1.31 \\\hline
SRoBERTa-Base + \texttt{[CLS]} + Cosine          &71.48  &77.73  &72.57  &77.79  &75.02  &72.76   &77.33   \\
SRoBERTa-Base + Max + Cosine          &70.60  &74.54  &71.73  &77.14  &73.79  &72.46   &76.27   \\
SRoBERTa-Base + Mean + Cosine       & \textbf{71.52}  & 75.47  & 72.59  & 78.37  & 73.81  & 74.38  & 77.8   \\
SRoBERTa-Base+ Latte-Mix + Cosine  & 70.42$\pm$0.27 & 76.89$\pm$0.47 & 75.69$\pm$0.31 & 81.03$\pm$0.16 & 74.99$\pm$0.51 & 74.56$\pm$0.1  & 78.79$\pm$0.19 \\
SRoBERTa-Base + Latte-Mix + JS  & 71.15$\pm$0.14 & \textbf{78.59$\pm$0.2}  & \textbf{76.81$\pm$0.15} & \textbf{82.08$\pm$0.15} & \textbf{76.49$\pm$0.34} & \textbf{75.6$\pm$0.02}  & \textbf{79.95$\pm$0.09} \\
SRoBERTa-Base + Latte-Mix + L2  & 68.81$\pm$0.39 & 76.72$\pm$0.47 & 71.86$\pm$0.16 & 68.69$\pm$0.42 & 69.47$\pm$0.5  & 67.39$\pm$0.19 & 74.02$\pm$0.36 \\\hline
SRoBERTa-Large + \texttt{[CLS]} + Cosine          &71.49  &78.21  &71.76  &78.64  &74.47  &71.89   &77.03   \\
SRoBERTa-Large + Max + Cosine          &73.66  &80.04  &74.80  &81.09  &77.50  &73.80   &79.27   \\
SRoBERTa-Large + Mean + Cosine      & \textbf{73.91}  & 79.71  & 74.11  & 81.62  & 77.91  & 74.12  & 79.19  \\
SRoBERTa-Large+ Latte-Mix + Cosine & 72.65$\pm$0.28 & 80.71$\pm$0.12 & 77.14$\pm$0.45 & 82.53$\pm$0.22 & 78.3$\pm$0.22  & 75.22$\pm$0.52 & 80.06$\pm$0.24 \\
SRoBERTa-Large + Latte-Mix + JS & 72.9$\pm$0.22  & \textbf{82.67$\pm$0.1}  & \textbf{78.42$\pm$0.24} & \textbf{83.95$\pm$0.15} & \textbf{79.85$\pm$0.18} & \textbf{75.97$\pm$0.31} & \textbf{81.6$\pm$0.01} \\
SRoBERTa-Large + Latte-Mix + L2 & 70.25$\pm$0.09 & 79.61$\pm$0.46 & 73.0$\pm$0.35  & 68.74$\pm$1.44 & 71.63$\pm$0.53 & 66.85$\pm$0.89 & 73.43$\pm$0.28 \\ \bottomrule
  \end{tabular}
}
  \caption{Zero-shot spearman's rank correlation $\rho\times 100$ of the negative distance between sentence embeddings/distributions and the gold labels on STS tasks. For Sentence-BERT and Sentence-RoBERTa, outputs at the last layer are used as inputs for the token aggregation methods. STS12-STS16: SemEval 2012-2016, STSb: STSbenchmark, SICK-R: SICK relatedness dataset. SBERT: Sentence-BERT fintuned on the SNLI dataset. SRoBERTa: Sentence-RoBERTa fintuned on the SNLI dataset. Cosine: Cosine similarity. Mean: Mean pooling. Max: Max pooling. \texttt{[CLS]}: the embedding for the \texttt{[CLS]} token. Latte-Mix: Latent mixtures from the Categorical VAE model. Cosine: Cosine similarity. L2: $\ell_2$ distance. JS: Jason-Shannon Divergence.}
  \label{tbl:sbert_sts_all}
\end{table*}

%% file: tables/bert_base_layers_sts_all.tex
\begin{table*}
  \centering
\resizebox{18cm}{!}{
  \begin{tabular}{l@{\hskip .15in}l@{\hskip 0.15in}l@{\hskip 0.15in}l@{\hskip 0.15in}l@{\hskip 0.15in}l@{\hskip 0.15in}l@{\hskip 0.15in}l@{\hskip 0.15in}}
\toprule

Methods &STS12 &STS13 &STS14 &STS15 &STS16 &SICK-R &STSb \\\midrule
BERT-Base, Layer 1 + Mean + Cosine   & \textbf{50.78}  & 69.12  & \textbf{54.22}  & \textbf{71.24}  & \textbf{64.36}  & \textbf{61.78}  & \textbf{58.14}  \\
BERT-Base, Layer 2 + Mean + Cosine   & 49.4   & \textbf{69.13}  & 52.01  & 68.7   & 62.19  & 61.65  & 56.07  \\
BERT-Base, Layer 3 + Mean + Cosine   & 46.25  & 68.68  & 49.72  & 66.31  & 61.0   & 60.55  & 54.77  \\
BERT-Base, Layer 4 + Mean + Cosine   & 42.47  & 67.9   & 47.32  & 63.5   & 58.85  & 58.85  & 52.12  \\
BERT-Base, Layer 5 + Mean + Cosine   & 41.24  & 67.57  & 47.45  & 64.05  & 60.59  & 58.52  & 52.18  \\
BERT-Base, Layer 6 + Mean + Cosine   & 38.58  & 66.59  & 46.57  & 63.72  & 61.54  & 57.76  & 51.22  \\
BERT-Base, Layer 7 + Mean + Cosine   & 37.66  & 66.02  & 47.33  & 64.34  & 62.41  & 57.73  & 51.29  \\
BERT-Base, Layer 8 + Mean + Cosine   & 36.84  & 65.91  & 46.71  & 64.27  & 61.39  & 58.5   & 49.56  \\
BERT-Base, Layer 9 + Mean + Cosine   & 33.4   & 62.85  & 46.1   & 63.28  & 60.03  & 57.45  & 45.49 \\
BERT-Base, Layer 10 + Mean + Cosine & 35.69  & 65.16  & 49.18  & 64.43  & 62.27  & 58.01  & 46.23  \\
BERT-Base, Layer 11 + Mean + Cosine & 36.98  & 64.39  & 49.08  & 63.55  & 61.53  & 58.47  & 47.39  \\
BERT-Base, Layer 12 + Mean + Cosine & 30.88  & 65.78  & 47.74  & 60.29  & 63.73  & 58.22  & 47.29  \\\bottomrule
BERT-Base, Layer 1 + Latte-Mix + Cosine  & \textbf{50.77$\pm$1.48} & \textbf{65.94$\pm$0.84} & \textbf{58.13$\pm$1.4}  & \textbf{74.01$\pm$1.13} & \textbf{63.54$\pm$1.18} & \textbf{61.33$\pm$0.44} & \textbf{61.47$\pm$1.44} \\
BERT-Base, Layer 2 + Latte-Mix + Cosine  & 47.54$\pm$0.73 & 65.38$\pm$1.01 & 55.12$\pm$1.31 & 71.29$\pm$1.54 & 60.64$\pm$1.29 & 60.93$\pm$0.46 & 58.88$\pm$1.14 \\
BERT-Base, Layer 3 + Latte-Mix + Cosine  & 44.6$\pm$0.96  & 63.95$\pm$0.42 & 52.3$\pm$0.53  & 69.01$\pm$0.45 & 58.27$\pm$0.34 & 58.57$\pm$0.19 & 56.97$\pm$0.18 \\
BERT-Base, Layer 4 + Latte-Mix + Cosine  & 43.4$\pm$0.82  & 64.17$\pm$0.34 & 51.25$\pm$0.49 & 67.59$\pm$0.89 & 57.38$\pm$0.52 & 57.13$\pm$0.34 & 55.91$\pm$0.67 \\
BERT-Base, Layer 5 + Latte-Mix + Cosine  & 44.64$\pm$1.5  & 63.67$\pm$0.83 & 51.55$\pm$0.34 & 68.67$\pm$0.34 & 58.57$\pm$0.19 & 57.52$\pm$0.21 & 56.86$\pm$1.45 \\
BERT-Base, Layer 6 + Latte-Mix + Cosine  & 44.48$\pm$1.15 & 63.26$\pm$0.22 & 50.2$\pm$0.53  & 68.89$\pm$1.08 & 59.51$\pm$0.4  & 56.77$\pm$0.56 & 56.94$\pm$0.85 \\
BERT-Base, Layer 7 + Latte-Mix + Cosine  & 41.93$\pm$1.1  & 62.99$\pm$0.54 & 49.09$\pm$0.25 & 67.23$\pm$1.11 & 59.36$\pm$0.28 & 56.71$\pm$0.34 & 55.67$\pm$0.85 \\
BERT-Base, Layer 8 + Latte-Mix + Cosine  & 41.8$\pm$0.7   & 64.82$\pm$0.64 & 49.32$\pm$0.42 & 67.2$\pm$0.74  & 58.59$\pm$0.24 & 58.5$\pm$0.58  & 55.22$\pm$0.74 \\
BERT-Base, Layer 9 + Latte-Mix + Cosine  & 41.92$\pm$1.66 & 63.66$\pm$0.77 & 49.41$\pm$0.43 & 67.84$\pm$0.71 & 59.61$\pm$0.46 & 59.23$\pm$0.25 & 55.16$\pm$1.49\\
BERT-Base, Layer 10 + Latte-Mix + Cosine & 43.86$\pm$0.29 & 66.58$\pm$0.2  & 52.46$\pm$0.29 & 68.99$\pm$0.7  & 62.4$\pm$0.58  & 60.46$\pm$0.13 & 56.91$\pm$0.85 \\
BERT-Base, Layer 11 + Latte-Mix + Cosine & 44.3$\pm$0.98  & 66.62$\pm$0.49 & 52.85$\pm$0.11 & 67.7$\pm$0.45  & 62.96$\pm$0.89 & 60.76$\pm$0.13 & 56.83$\pm$0.77 \\
BERT-Base, Layer 12 + Latte-Mix + Cosine & 43.6$\pm$1.26  & 67.47$\pm$0.58 & 53.56$\pm$0.53 & 68.06$\pm$1.3  & 63.02$\pm$1.62 & 60.7$\pm$0.77  & 56.92$\pm$1.05 \\\bottomrule
  \end{tabular}
}
  \caption{Zero-shot spearman's rank correlation $\rho\times 100$ of the negative distance between sentence embeddings/distributions and the gold labels on STS tasks. Results of using different layers of the BERT-Base model as inputs with mean pooling and Latte-Mix are reported. STS12-STS16: SemEval 2012-2016, STSb: STSbenchmark, SICK-R: SICK relatedness dataset. Cosine: Cosine similarity. Mean: Mean pooling. Layer N: using the Nth layer of the pretrained model as word embeddings.}
  \label{tbl:base_layer_sts_all}
\end{table*}

%% file: tables/bert_base_layers_web_all.tex
\begin{table*}
  \centering
\resizebox{15cm}{!}{
  \begin{tabular}{l@{\hskip .15in}c@{\hskip 0.15in}c@{\hskip 0.15in}c@{\hskip 0.15in}c@{\hskip 0.15in}c@{\hskip 0.15in}c@{\hskip 0.15in}c@{\hskip 0.15in}c@{\hskip 0.15in}c@{\hskip 0.15in}}
\toprule

Methods & SimLex999 & MEN  & WS353 & RW   & Google & MSR  & SemEval2012\_2 & BLESS & AP \\\midrule

BERT-Base, Layer 1        & \textbf{34.85}     & \textbf{27.41} & \textbf{43.32} & \textbf{21.09} & 26.52  & \textbf{40.49} & \textbf{14.42}          & \textbf{38.5}  & \textbf{33.83} \\

BERT-Base, Layer 2        & 33.59     & 25.38 & 40.27 & 18.24 & \textbf{26.69}  & 40.05 & 16.17          & 38.0  & 33.58 \\

BERT-Base, Layer 3        & 34.01     & 24.42 & 41.69 & 18.57 & 25.1   & 37.94 & 15.04          & 39.0  & 33.58 \\

BERT-Base, Layer 4        & 32.52     & 22.48 & 37.81 & 18.22 & 24.33  & 37.21 & 15.38          & 39.5  & 34.58 \\

BERT-Base, Layer 5        & 31.47     & 21.98 & 37.25 & 17.23 & 22.69  & 35.45 & 16.22          & 38.5  & 33.33 \\

BERT-Base, Layer 6        & 30.83     & 21.03 & 37.09 & 16.46 & 22.43  & 35.1  & 16.51          & 39.5  & 33.08 \\

BERT-Base, Layer 7        & 29.88     & 21.27 & 37.24 & 15.88 & 21.72  & 33.89 & 15.51          & 40.0  & 32.59 \\

BERT-Base, Layer 8        & 29.04     & 22.63 & 35.88 & 15.14 & 21.15  & 32.89 & 15.53          & 39.0  & 33.33 \\

BERT-Base, Layer 9        & 27.47     & 22.38 & 35.84 & 15.55 & 20.46  & 30.79 & 14.9           & 39.5  & 33.33 \\

BERT-Base, Layer 10       & 26.48     & 21.51 & 32.75 & 15.15 & 19.49  & 27.78 & 13.88          & 38.5  & 33.58 \\

BERT-Base, Layer 11       & 25.92     & 19.34 & 29.64 & 14.78 & 19.43  & 27.81 & 13.79          & 38.5  & 31.34 \\

BERT-Base, Layer 12       & 24.03     & 22.08 & 32.58 & 12.89 & 19.29  & 28.22 & 13.49          & 38.0  & 29.6  \\\bottomrule

  \end{tabular}
}
  \caption{The performance of using the first principal component of the BERT-Base model's different layers as the static embeddings on various tasks in word embedding benchmark. The evaluation metrics are different for each task, but the higher score means the better performance for all tasks. The principle component is computed using singular value decomposition on the STS dataset. Layer N: using the Nth layer of the pretrained model as word embeddings.}
  \label{tbl:base_layers_web_all}
\end{table*}

%% file: tables/bert_large_layers_sts_all.tex
\begin{table*}
  \centering
\resizebox{18cm}{!}{
  \begin{tabular}{l@{\hskip .15in}l@{\hskip 0.15in}l@{\hskip 0.15in}l@{\hskip 0.15in}l@{\hskip 0.15in}l@{\hskip 0.15in}l@{\hskip 0.15in}l@{\hskip 0.15in}}
\toprule

Methods &STS12 &STS13 &STS14 &STS15 &STS16 &SICK-R &STSb \\\midrule
BERT-Large, Layer 1 + Mean + Cosine   & \textbf{53.29}  & 69.4   & 55.51  & 71.18  & 63.55  & 59.98  & 58.73  \\
BERT-Large, Layer 2 + Mean + Cosine   & 53.01  & 69.61  & 55.49  & 70.96  & 63.7   & 60.23  & 59.04  \\
BERT-Large, Layer 3 + Mean + Cosine   & 51.63  & 69.28  & 54.45  & 68.94  & 63.1   & 60.66  & 57.85  \\
BERT-Large, Layer 4 + Mean + Cosine   & 50.91  & 69.11  & 55.18  & 69.89  & 64.38  & 60.92  & 58.28  \\
BERT-Large, Layer 5 + Mean + Cosine   & 50.11  & 69.13  & 56.46  & 71.27  & 65.72  & 61.63  & 59.77  \\
BERT-Large, Layer 6 + Mean + Cosine  & 48.62  & 68.47  & \textbf{56.89}  & \textbf{71.94}  & \textbf{67.71}  & \textbf{62.06}  & \textbf{61.07}  \\
BERT-Large, Layer 7 + Mean + Cosine   & 45.59  & 69.91  & 53.28  & 68.61  & 63.45  & 60.84  & 56.75  \\
BERT-Large, Layer 8 + Mean + Cosine   & 45.76  & 69.89  & 51.01  & 66.35  & 62.69  & 60.87  & 56.15  \\
BERT-Large, Layer 9 + Mean + Cosine   & 44.8   & \textbf{70.83}  & 49.97  & 65.27  & 62.61  & 60.69  & 55.88\\
BERT-Large, Layer 10 + Mean + Cosine & 44.14  & 70.58  & 49.18  & 65.53  & 61.61  & 60.31  & 55.15  \\
BERT-Large, Layer 11 + Mean + Cosine & 39.73  & 69.17  & 45.33  & 62.3   & 58.45  & 59.13  & 51.37  \\
BERT-Large, Layer 12 + Mean + Cosine & 33.19 & 66.91 & 43.67 & 57.49 & 58.83 & 56.46 & 48.52 \\
BERT-Large, Layer 13 + Mean + Cosine & 34.61  & 67.28  & 42.96  & 61.63  & 58.31  & 57.48  & 49.04  \\
BERT-Large, Layer 14 + Mean + Cosine & 34.41  & 67.09  & 46.33  & 64.81  & 62.11  & 57.84  & 51.4   \\
BERT-Large, Layer 15 + Mean + Cosine & 36.75  & 67.21  & 47.72  & 65.61  & 64.31  & 59.29  & 53.67  \\
BERT-Large, Layer 16 + Mean + Cosine & 33.46  & 66.8   & 45.08  & 62.97  & 61.88  & 58.08  & 48.35  \\
BERT-Large, Layer 17 + Mean + Cosine & 32.35  & 64.49  & 45.45  & 63.87  & 60.6   & 58.23  & 45.98  \\
BERT-Large, Layer 18 + Mean + Cosine & 33.0   & 64.97  & 46.42  & 64.46  & 60.67  & 57.28  & 45.93  \\
BERT-Large, Layer 19 + Mean + Cosine & 33.6   & 64.66  & 46.66  & 64.16  & 60.76  & 57.64  & 45.65  \\
BERT-Large, Layer 20 + Mean + Cosine & 34.66  & 64.14  & 47.37  & 63.59  & 60.94  & 56.96  & 46.85  \\
BERT-Large, Layer 21 + Mean + Cosine & 34.83  & 66.66  & 48.6   & 60.89  & 60.93  & 55.47  & 47.73  \\
BERT-Large, Layer 22 + Mean + Cosine & 35.55  & 67.45  & 48.58  & 59.24  & 61.38  & 53.85  & 48.8   \\
BERT-Large, Layer 23 + Mean + Cosine & 35.72  & 67.22  & 47.9   & 56.3   & 60.93  & 53.18  & 49.84  \\
BERT-Large, Layer 24 + Mean + Cosine & 27.69  & 63.97  & 44.48  & 51.67  & 61.85  & 53.85  & 47.0   \\\bottomrule
BERT-Large, Layer 1 + Latte-Mix + Cosine  & \textbf{53.53$\pm$0.49} & 68.58$\pm$0.14 & 58.24$\pm$0.42 & 73.22$\pm$2.48 & 64.62$\pm$0.21 & 60.48$\pm$0.62 & \textbf{61.62$\pm$0.39} \\
BERT-Large, Layer 2 + Latte-Mix + Cosine  & 51.9$\pm$1.25  & 67.85$\pm$0.93 & 55.98$\pm$0.71 & 71.64$\pm$0.87 & 62.79$\pm$1.11 & 60.72$\pm$0.56 & 59.83$\pm$0.86 \\
BERT-Large, Layer 3 + Latte-Mix + Cosine  & 50.4$\pm$0.37  & 67.57$\pm$0.94 & 54.8$\pm$0.48  & 69.81$\pm$0.12 & 61.82$\pm$1.48 & 60.54$\pm$0.29 & 58.36$\pm$0.61 \\
BERT-Large, Layer 4 + Latte-Mix + Cosine  & 49.59$\pm$0.47 & 67.55$\pm$0.81 & 55.53$\pm$0.45 & 70.21$\pm$0.57 & 62.38$\pm$0.36 & 60.67$\pm$0.33 & 58.56$\pm$0.32 \\
BERT-Large, Layer 5 + Latte-Mix + Cosine  & 48.89$\pm$0.47 & 67.27$\pm$0.04 & 56.03$\pm$0.54 & 71.09$\pm$0.37 & 63.36$\pm$2.59 & 61.01$\pm$0.73 & 59.48$\pm$0.76 \\
BERT-Large, Layer 6 + Latte-Mix + Cosine  & 48.04$\pm$0.66 & 67.52$\pm$1.05 & \textbf{56.75$\pm$0.26} & \textbf{71.86$\pm$0.48} & \textbf{64.81$\pm$0.21} & \textbf{61.36$\pm$0.81} & 60.8$\pm$0.32  \\
BERT-Large, Layer 7 + Latte-Mix + Cosine  & 45.99$\pm$0.51 & 68.66$\pm$0.39 & 54.28$\pm$0.17 & 69.82$\pm$0.46 & 62.65$\pm$0.95 & 60.39$\pm$0.58 & 58.06$\pm$0.44 \\
BERT-Large, Layer 8 + Latte-Mix + Cosine  & 46.27$\pm$0.57 & 68.84$\pm$0.2  & 53.04$\pm$0.24 & 68.27$\pm$0.22 & 62.23$\pm$0.66 & 60.33$\pm$0.6  & 57.57$\pm$0.59 \\
BERT-Large, Layer 9 + Latte-Mix + Cosine & 45.45$\pm$0.75 & \textbf{69.05$\pm$0.06} & 51.79$\pm$0.03 & 67.38$\pm$0.43 & 61.88$\pm$0.02 & 60.1$\pm$0.67  & 57.09$\pm$0.36 \\
BERT-Large, Layer 10 + Latte-Mix + Cosine & 44.46$\pm$0.4  & 69.14$\pm$0.66 & 51.2$\pm$0.21  & 66.84$\pm$0.47 & 61.03$\pm$0.82 & 59.62$\pm$0.75 & 56.82$\pm$0.85 \\
BERT-Large, Layer 11 + Latte-Mix + Cosine & 41.28$\pm$0.72 & 68.04$\pm$0.27 & 48.32$\pm$0.28 & 64.34$\pm$0.34 & 58.43$\pm$0.25 & 58.34$\pm$0.95 & 53.98$\pm$0.87 \\
BERT-Large, Layer 12 + Latte-Mix + Cosine & 37.32$\pm$0.44 & 66.62$\pm$0.71 & 47.25$\pm$0.08 & 61.02$\pm$0.29 & 59.59$\pm$0.27 & 56.59$\pm$0.47 & 52.48$\pm$0.54 \\
BERT-Large, Layer 13 + Latte-Mix + Cosine & 37.73$\pm$0.43 & 66.37$\pm$0.03 & 46.37$\pm$0.75 & 64.18$\pm$0.8  & 58.18$\pm$0.33 & 56.95$\pm$0.67 & 51.87$\pm$0.1  \\
BERT-Large, Layer 14 + Latte-Mix + Cosine & 37.69$\pm$0.59 & 66.11$\pm$0.19 & 48.38$\pm$0.28 & 66.46$\pm$0.84 & 61.32$\pm$0.9  & 57.45$\pm$0.45 & 53.49$\pm$0.36 \\
BERT-Large, Layer 15 + Latte-Mix + Cosine & 39.71$\pm$0.36 & 66.18$\pm$0.13 & 49.08$\pm$0.5  & 66.83$\pm$0.4  & 62.99$\pm$1.47 & 58.96$\pm$0.4  & 55.55$\pm$0.12 \\
BERT-Large, Layer 16 + Latte-Mix + Cosine & 37.61$\pm$0.55 & 65.89$\pm$0.15 & 46.93$\pm$0.19 & 64.47$\pm$0.71 & 61.42$\pm$0.5  & 58.24$\pm$0.23 & 51.41$\pm$0.39 \\
BERT-Large, Layer 17 + Latte-Mix + Cosine & 37.07$\pm$0.19 & 64.67$\pm$0.29 & 47.32$\pm$0.06 & 65.11$\pm$0.41 & 60.92$\pm$0.84 & 58.65$\pm$0.48 & 49.78$\pm$0.17 \\
BERT-Large, Layer 18 + Latte-Mix + Cosine & 37.8$\pm$0.25  & 64.86$\pm$0.4  & 47.69$\pm$0.41 & 65.83$\pm$0.51 & 61.48$\pm$0.91 & 58.49$\pm$0.33 & 50.09$\pm$0.56 \\
BERT-Large, Layer 19 + Latte-Mix + Cosine & 37.92$\pm$0.76 & 65.44$\pm$0.89 & 48.57$\pm$0.1  & 66.11$\pm$0.15 & 61.39$\pm$0.74 & 58.99$\pm$0.48 & 49.89$\pm$0.67 \\
BERT-Large, Layer 20 + Latte-Mix + Cosine & 38.75$\pm$0.5  & 65.11$\pm$0.18 & 49.07$\pm$0.9  & 65.65$\pm$0.27 & 61.95$\pm$0.16 & 58.28$\pm$0.49 & 50.85$\pm$0.41 \\
BERT-Large, Layer 21 + Latte-Mix + Cosine & 39.3$\pm$0.91  & 67.07$\pm$0.64 & 50.32$\pm$0.88 & 63.47$\pm$0.85 & 61.87$\pm$0.03 & 57.21$\pm$0.92 & 51.89$\pm$0.57 \\
BERT-Large, Layer 22 + Latte-Mix + Cosine & 39.55$\pm$0.39 & 68.04$\pm$0.79 & 50.63$\pm$0.26 & 61.52$\pm$0.61 & 62.15$\pm$0.96 & 55.6$\pm$0.94  & 52.65$\pm$0.22 \\
BERT-Large, Layer 23 + Latte-Mix + Cosine & 39.34$\pm$0.97 & 67.48$\pm$0.4  & 50.16$\pm$0.49 & 59.08$\pm$0.06 & 62.51$\pm$0.83 & 54.96$\pm$0.96 & 53.79$\pm$0.37 \\
BERT-Large, Layer 24 + Latte-Mix + Cosine & 34.06$\pm$0.98 & 65.45$\pm$0.64 & 48.43$\pm$0.32 & 56.8$\pm$0.63  & 63.62$\pm$0.95 & 55.43$\pm$0.73 & 51.88$\pm$0.34 \\ \bottomrule
  \end{tabular}
}
  \caption{Zero-shot spearman's rank correlation $\rho\times 100$ of the negative distance between sentence embeddings/distributions and the gold labels on STS tasks. Results of using different layers of the BERT-Large model as inputs with mean pooling and Latte-Mix are reported. STS12-STS16: SemEval 2012-2016, STSb: STSbenchmark, SICK-R: SICK relatedness dataset. Cosine: Cosine similarity. Mean: Mean pooling. Layer N: using the Nth layer of the pretrained model as word embeddings.}
  \label{tbl:large_layer_sts_all}
\end{table*}

%% file: tables/bert_large_layers_web_all.tex
\begin{table*}
  \centering
\resizebox{15cm}{!}{
  \begin{tabular}{l@{\hskip .15in}c@{\hskip 0.15in}c@{\hskip 0.15in}c@{\hskip 0.15in}c@{\hskip 0.15in}c@{\hskip 0.15in}c@{\hskip 0.15in}c@{\hskip 0.15in}c@{\hskip 0.15in}c@{\hskip 0.15in}}
\toprule

Methods & SimLex999 & MEN  & WS353 & RW   & Google & MSR  & SemEval2012\_2 & BLESS & AP \\\midrule
 
BERT-Large, Layer 1  & 29.43     & 19.67 & 41.35 & 15.89 & 25.89  & \textbf{39.9}  & 15.17          & 38.0  & 32.84 \\
 
BERT-Large, Layer 2  & 27.89     & 18.75 & 40.6  & 15.41 & 25.95  & 39.39 & \textbf{15.48}          & 38.0  & 32.09 \\
 
BERT-Large, Layer 3  & 28.07     & 18.94 & 39.41 & 15.76 & \textbf{26.22}  & 39.41 & 14.93          & 38.0  & \textbf{34.08} \\
 
BERT-Large, Layer 4  & 29.02     & 19.65 & 41.14 & 16.69 & 25.62  & 38.06 & 16.11          & 37.0  & 33.33 \\
 
BERT-Large, Layer 5  & 30.82     & 21.59 & 42.47 & \textbf{17.85} & 25.26  & 37.44 & 15.12          & 37.0  & 32.34 \\
 
BERT-Large, Layer 6  & \textbf{31.02}     & \textbf{22.67} & \textbf{43.57} & 16.93 & 24.63  & 36.44 & 15.2           & 37.5  & 33.08 \\
 
BERT-Large, Layer 7  & 29.29     & 19.16 & 39.78 & 15.67 & 23.65  & 36.94 & 13.51          & \textbf{39.5}  & 33.08 \\
  
BERT-Large, Layer 8  & 28.4      & 19.11 & 38.98 & 15.36 & 23.79  & 37.05 & 14.46          & 37.5  & 31.59 \\
 
BERT-Large, Layer 9  & 29.06     & 18.79 & 37.57 & 14.62 & 23.43  & 36.49 & 14.47          & 37.5  & 33.58 \\
 
BERT-Large, Layer 10 & 28.83     & 17.83 & 36.62 & 14.37 & 22.61  & 36.26 & 13.97          & 36.5  & 32.84 \\

BERT-Large, Layer 11 & 27.68     & 16.66 & 36.12 & 14.38 & 22.51  & 36.16 & 14.89          & 38.0  & 34.08 \\
 
BERT-Large, Layer 12 & 26.37     & 16.58 & 36.63 & 14.52 & 21.19  & 34.02 & 13.7           & 37.5  & 32.09 \\
 
BERT-Large, Layer 13 & 26.27     & 16.43 & 36.54 & 14.53 & 21.1   & 33.25 & 14.17          & 37.5  & 31.59 \\

BERT-Large, Layer 14 & 26.76     & 18.43 & 37.81 & 14.4  & 21.15  & 31.49 & 13.41          & 38.5  & 32.09 \\
 
BERT-Large, Layer 15 & 28.1      & 20.6  & 40.13 & 13.96 & 21.14  & 30.94 & 13.22          & 37.0  & 33.58 \\
 
BERT-Large, Layer 16 & 27.2      & 18.89 & 38.07 & 13.54 & 20.68  & 30.02 & 12.93          & 38.0  & 32.34 \\
 
BERT-Large, Layer 17 & 26.39     & 18.92 & 37.31 & 13.5  & 20.04  & 27.95 & 12.68          & 37.5  & 33.33 \\
 
BERT-Large, Layer 18 & 26.97     & 19.69 & 37.9  & 13.08 & 19.26  & 26.59 & 12.48          & 38.5  & 33.33 \\
 
BERT-Large, Layer 19 & 27.47     & 19.85 & 38.45 & 13.51 & 19.56  & 25.98 & 13.94          & 38.5  & 32.84 \\
 
BERT-Large, Layer 20 & 27.32     & 19.41 & 38.47 & 13.55 & 19.46  & 24.82 & 13.03          & 38.5  & 33.83 \\
 
BERT-Large, Layer 21 & 27.44     & 17.63 & 37.98 & 13.29 & 18.8   & 26.0  & 12.33          & 38.0  & 33.33 \\

BERT-Large, Layer 22 & 25.94     & 16.2  & 36.75 & 12.93 & 18.73  & 26.19 & 12.18          & 37.5  & 32.59 \\
 
BERT-Large, Layer 23 & 23.66     & 14.0  & 34.27 & 12.91 & 18.72  & 26.04 & 12.13          & 37.5  & 31.34 \\
 
BERT-Large, Layer 24 & 22.64     & 14.97 & 32.19 & 12.61 & 18.95  & 27.31 & 10.54          & 36.0  & 30.1  \\\bottomrule

\end{tabular}
}
  \caption{The performance of using the first principal component of the BERT-Large model's different layers as the static embeddings on various tasks in word embedding benchmark. The evaluation metrics are different for each task, but the higher score means the better performance for all tasks. The principle component is computed using singular value decomposition on the STS dataset. Layer N: using the Nth layer of the pretrained model as word embeddings.}
  \label{tbl:large_layers_web_all}
\end{table*}